\documentclass{article} 
\usepackage{iclr2024,times}
\iclrfinalcopy 

\usepackage[utf8]{inputenc} 
\usepackage[T1]{fontenc}    
\usepackage{url}            
\usepackage{booktabs}       
\usepackage{amsfonts}       
\usepackage{nicefrac}       
\usepackage{microtype}      
\usepackage{xcolor}         
\usepackage[pagebackref=true,breaklinks=true,colorlinks,urlcolor=black,bookmarks=false]{hyperref}
\hypersetup{citecolor=[RGB]{0,139,69}}

\usepackage{subfig}
\usepackage{graphicx}
\usepackage{caption}
\usepackage{amsmath}
\usepackage{amssymb}
\usepackage{amsthm}
\usepackage{enumerate}
\usepackage{multirow}
\usepackage{adjustbox}
\usepackage{epsfig}
\usepackage{ulem}
\usepackage{colortbl}
\definecolor{mygray}{gray}{0.9}
\newcolumntype{a}{>{\columncolor{mygray}}c|c}
\usepackage{marvosym}

\usepackage[capitalize]{cleveref}
\crefname{section}{Sec.}{Secs.}
\Crefname{section}{Section}{Sections}
\Crefname{table}{Table}{Tables}
\crefname{table}{Tab.}{Tabs.}

\begin{document}

\title{A Study of Unsupervised Evaluation Metrics for Practical and Automatic Domain Adaptation\vspace{-3mm}}

\author{
Minghao Chen$^1$\quad
Zepeng Gao$^2$\quad
Shuai Zhao$^{3,5}$\quad
Qibo Qiu$^4$\quad
Wenxiao Wang$^2$\quad \\
\textbf{Binbin Lin$^{2}\textsuperscript{\Letter}$ \quad
Xiaofei He$^1$}
\smallskip
\\
$^1$State Key Lab of CAD\&CG, College of Computer Science, Zhejiang University
\\
$^2$School of Software Technology, Zhejiang University \\
$^3$ReLER Lab, CCAI, Zhejiang University \quad
$^4$Zhejiang Lab \quad $^5$Baidu Inc.
\\
\tt\small \{minghaochen01,zhaoshuaimcc\}@gmail.com
\quad binbinlin@zju.edu.cn  
}

\maketitle

\vspace{-5mm}

\begin{abstract}
  Unsupervised domain adaptation (UDA) methods facilitate the transfer of models to target domains without labels. However, these methods necessitate a labeled target validation set for hyper-parameter tuning and model selection. In this paper, we aim to find an evaluation metric capable of assessing the quality of a transferred model without access to target validation labels. We begin with the metric based on mutual information of the model prediction. Through empirical analysis, we identify three prevalent issues with this metric: 1) It does not account for the source structure. 2) It can be easily attacked. 3) It fails to detect negative transfer caused by the over-alignment of source and target features. To address the first two issues, we incorporate source accuracy into the metric and employ a new MLP classifier that is held out during training, significantly improving the result. To tackle the final issue, we integrate this enhanced metric with data augmentation, resulting in a novel unsupervised UDA metric called the Augmentation Consistency Metric (ACM). Additionally, we empirically demonstrate the shortcomings of previous experiment settings and conduct large-scale experiments to validate the effectiveness of our proposed metric. Furthermore, we leverage our metric to automatically search for the optimal set of hyper-parameters, achieving superior performance comparable to manually tuned sets across four common benchmarks.
\end{abstract}
\vspace{-2mm}

\section{Introduction}

Deep neural networks, when trained on extensive datasets, have demonstrated exceptional performance across various computer vision tasks such as classification~\cite{ConvNeXt, CLIP}, object detection~\cite{DETR, DINO}, and semantic segmentation~\cite{Deeplabv3+, SegFormer}. Some even exhibit remarkable generalization to unseen domains~\cite{ViLD, SEEM}. But performance in specific domains can always be enhanced through fine-tuning with labels. The challenge arises in real-world applications where manually labeling ample data for fine-tuning is both costly and impractical. Consider a household robot equipped with a vision system trained on vast vision datasets~\cite{RobotLifelong}. When introduced to a new home, it's anticipated that the robot would automatically adapt to this new environment by collecting images from the house. Yet, expecting homeowners to label these images is both burdensome and unrealistic.

Unsupervised domain adaptation (UDA) emerged as a solution to this problem. Over recent years, a plethora of UDA methods~\citep{CDAN,MCD,MDD,MCC,Proto} have been developed to facilitate transfer to label-free target domains. While accuracy in these domains has seen improvement, it often hinges on meticulous tuning using a labeled validation set from the target domain, which can be resource-intensive. Referring back to the household robot example, creating validation sets would necessitate precise labels from homeowners, which hinders the fully automatic adaptation. One could pre-test UDA method hyper-parameters in sample homes before actual deployment, selecting parameters that perform well across these homes. However, as UDA methods tend to be hyper-parameter sensitive, different domains might demand distinct hyper-parameter configurations. It is challenging to finalize an optimal set before deployment.

DEV~\cite{DEV} first proposed a general UDA evaluation metric, which uses the importance-weighted validation method~\cite{IWV} with a variance control term. Later, SND~\cite{SND} suggested that a good transfer model should have a compact neighborhood for each target feature and introduced the soft neighborhood density metric. However, upon more comprehensive and detailed experiments with UDA evaluation metrics, we discovered previous evaluation metrics often failed to select suitable models in most scenarios, as shown in Tab~\ref{tab:conclusion-dataset}. 
This is because their metrics are based on assumptions that do not always hold true in a wide range of scenarios. More related discussions refer to Related Works.

\begin{table*}[t]
\vspace{-6mm}
    \centering
    \resizebox{1\textwidth}{!}{    
        \begin{tabular}{clc}
            \toprule
            Metrics & Ar $\to$ Cl Ar $\to$ Pr Ar $\to$ Rw Cl $\to$ Ar Cl $\to$ Pr Cl $\to$ Rw\,Pr $\to$ Ar Pr $\to$ Cl Pr $\to$ Rw Rw $\to$ Ar Rw $\to$ Cl\,Rw $\to$ Pr  & Avg    \\
            \hline
            DEV~\cite{DEV} & \quad 4.50 \qquad 2.62\qquad 1.98\qquad1.92\qquad 0.53\qquad 9.25\qquad 1.78\qquad 16.9\qquad 8.10       \,\,\,\qquad 0.82       \,\qquad 0.61      \,\qquad \textbf{0.00} & 4.07\\
            SND~\cite{SND} & \quad 16.6 \qquad 2.62 \qquad 1.91\qquad22.6\qquad 24.4\qquad 19.3\qquad 14.1\qquad 16.9\qquad \textbf{0.00}       \,\,\,\qquad 0.68       \,\qquad 16.9      \,\qquad 0.37 & 11.4\\
            \hline
            ACM (ours) & \quad \textbf{1.53} \qquad \textbf{0.00} \qquad \textbf{0.00}\qquad\textbf{0.68}\qquad \textbf{0.00}\qquad \textbf{1.07}\qquad \textbf{0.00}\qquad \textbf{0.00}\qquad \textbf{0.00}       \,\,\,\qquad \textbf{0.00}       \,\qquad \textbf{0.00}      \,\qquad \textbf{0.00} & \textbf{0.67}\\
            \bottomrule
    \end{tabular}}
    \caption{In all 12 transfer tasks from the OfficeHome dataset, we employ CDAN~\cite{CDAN} as the training method. The hyper-parameter space is defined as {trade-off=\{0.1,0.2,0.3,0.5,1.0,2.0,3.0\}}. We show the deviations between the optimal model determined by metrics and the true best target accuracy. In SND paper~\cite{SND} they only presented results for Ar $\to$ Pr and Rw $\to$ Ar.}
    \label{tab:conclusion-dataset}
    \vspace{-5mm}
\end{table*}

This realization led us to rethink and reassess what a robust UDA evaluation metric should be like. A robust UDA evaluation metric, as we define, should satisfy three principles: 1) Target Unsupervised. 2) Consistency with target accuracy in a wide range of scenarios. 3) Robustness: the metric should not be vulnerable to deliberately designed training methods and hyper-parameter sets. Previous works generally assume the first two principles; we augment this understanding by introducing the "Robustness" principle. This new perspective, inspired by the study of adversarial attacks in neural networks~\cite{Goodfellow2014ExplainingAH}, emphasizes the importance of designing metrics resistant to potential failure cases, thus leading to more robust evaluations.

Building upon this redefined understanding of a robust UDA evaluation metric, we turn our attention to a classic UDA algorithm, mutual information~\cite{C-Ent,MI}, used as an evaluation metric that measures the confidence (entropy) and diversity of the model on target samples. We meticulously dissect the metric to evaluate its adherence to the aforementioned three principles. Our investigation reveals three significant drawbacks with the metric: 1) unaware of the alignment between the prediction and the label, 2) easily attacked by designed training methods, 3) cannot detect negative transfer caused by the over-alignment between the source and target features. 
To address these issues, we first incorporate source accuracy into the metric to retain the source label structure. Then, we 
employ an additional MLP classifier held out during training to defend against attacks. We refer to this new metric as Inception Score Metric for UDA (ISM). Finally, we integrate the metric with data augmentation and propose Augmentation Consistency Metric (ACM) to evaluate models beyond features-level consideration.

We also establish new experimental settings for validating evaluation metrics, which contain sufficient datasets, training methods, and hyperparameter sets. In large-scale validation experiments, our evaluation metrics demonstrate high consistency with target accuracy in most cases.
The study of evaluation metrics also has the potential to advance AutoML~\cite{NAS,ENAS} research in the context of UDA. We employ simple hyperparameter optimization~\cite{Optuna} to illustrate this concept. Experiments show that the hyper-parameters automatically discovered by our metrics outperform manually tuned hyper-parameters for four popular UDA methods.





\section{Related Works}
\subsection{Unsupervised Domain Adaptation}
Unsupervised domain adaptation (UDA)~\cite{DAN} has been developed to save annotation effort during the transfer from the source domain to the target domain. Most UDA methods aim to reduce the divergence~\cite{BenDavidA,BenDavidH} between the source and target domains, e.g., Discrepancy-based UDA ~\cite{CAN,Deepcoral}, Domain Adversarial UDA ~\cite{DANN,CDAN,MCD,MDD}, self-supervised-based UDA ~\cite{Selfensembling,MCC}.
However, none has formally studied whether their methods can decide the best model or how to tune hyper-parameters without target labels. 

\subsection{Model Selection for UDA} 

Some previous papers~\cite{DEV,SND} are also interested in the unsupervised evaluation metric for UDA, also known as the model selection for UDA.

\textbf{Importance Weighted Validation:} In~\cite{CDAN}, they tune the trade-off parameter using importance-weighted cross-validation~\cite{IWV}. In the later work~\cite{DEV}, Deep Embedded Validation (DEV) is proposed to select models based on importance weights and control variates. However, their DEV metric requires overlap between the support sets of two domain distributions. In practice, it could easily collapse when no overlap exists between two domain distributions or the source error becomes 0.


\textbf{Entropy-based Metric:} Morerio et al.~\cite{C-Ent} used the predicted entropy of the target domain samples as the metric to tune the hyper-parameter. Although it is simple and convenient, it was pointed out that it cannot deal with blind confidence~\cite{SND}. SND~\cite{SND} proposes to use the density of the target domain sample domain as an evaluation metric to solve this problem. However, SND cannot solve the ``mode collapse'' problem where all target samples are mapped into one feature point.

\textbf{Other Metrics:} BPDA~\cite{Balancing} introduces a principle to balance the source supervised error and domain distance in a target error bound. However, their approach is limited to adjusting the trade-off hyper-parameter, leaving other hyper-parameters untouched. More recently, Dinu et al.~\cite{Aggregation} propose linear aggregations of vector-valued models to ensemble various models trained under different hyper-parameters. However, their resultant model aggregates all models across diverse hyper-parameters, demanding significantly more computational resources than a singular model, which is not practical for vision tasks.

\section{Methods}
\subsection{Problem Definition}
During the training of unsupervised domain adaptation, we have a labeled dataset sampled from the source domain, $\left\{\left(\boldsymbol{x}_i^s, y_i^s\right)\right\}_{i=1}^{n_s} \sim \mathcal{D}_s$ and an unlabeled dataset sampled from the target domain, $\left\{\boldsymbol{x}_j^t\right\}_{j=1}^{n_t} \sim \mathcal{D}_t$. We can apply off-the-shelf UDA methods~\cite{DANN,CDAN,MDD,MCC} to train a model $\boldsymbol{M}$ on these two training datasets. 
During the evaluation of UDA, we are given a labeled dataset from the source domain and an unlabeled dataset from the target domain, $\left\{(\tilde{\boldsymbol{x}}_i^s, \tilde{y}_i^s)\right\}_{i=1}^{\tilde{n}_s} \sim \mathcal{D}_s$ and $\left\{\tilde{\boldsymbol{x}}_j^t\right\}_{j=1}^{\tilde{n}_t} \sim \mathcal{D}_t$, which contain different samples from the training sets. Given a trained model $\boldsymbol{M}$
, an unsupervised evaluation metric for UDA should compute a score to reflect the classification accuracy of the model on the target domain $\mathcal{D}_t$, based on the evaluation sets and the model $\boldsymbol{M}$. As a common practice, the model is decomposed into a feature generator $\textbf{g}(\cdot)$ and a linear classifier $\textbf{f}(\cdot)$: $\boldsymbol{M}=\textbf{f}(\textbf{g}(\cdot))$. 

\subsection{Principles of A Robust Metric}
\label{sec:3.2}

We define the principles of a robust unsupervised evaluation metric for UDA as the following:

\textbf{1) Target Unsupervised:} The metric can only access the evaluation sets of UDA, $\left\{(\tilde{\boldsymbol{x}}_i^s, \tilde{y}_i^s)\right\}_{i=1}^{\tilde{n}_s}$ and $\left\{\tilde{\boldsymbol{x}}_j^t\right\}_{j=1}^{\tilde{n}_t}$, and the model $\boldsymbol{M}$. For versatility, the metric should be irrelevant to the training method.

\textbf{2) Consistency:} Given a bunch of models $\{\boldsymbol{M}_l\}_{l=1}^{n_m}$ trained with different UDA methods and different hyper-parameters, the metric score $\{\boldsymbol{S}_l\}_{l=1}^{n_m}$ should be consistent with the target classification accuracy $\{\boldsymbol{A}_l\}_{l=1}^{n_m}$ and this consistency holds for multiple UDA datasets. 

\textbf{3) Robustness:} The metric should maintain consistency when we deliberately design the training method and the hyper-parameter to attack the metric. Typically, if the metric can be transformed to a training loss for UDA, the metric score should still be consistent with the target accuracy when training with this loss. 

Our intuition is that a robust metric should reflect the target domain accuracy under various conditions. At the same time, a robust metric should not be vulnerable to attack, which avoids some methods of deliberately optimizing the metric and finding metric preferences. Just as the robustness of the neural network can be improved through the attack on the neural network~\cite{AdversarialExamples}, the analysis of the attack on the evaluation metric can help us construct a more robust evaluation metric.

We utilize two measurements to measure the degree of consistency between the metric score and target accuracy: 

\textbf{Pearson's correlation coefficient:}
\begin{align}
corr(\{\boldsymbol{S}_l\}_{l=1}^{n_m}, \{\boldsymbol{A}_l\}_{l=1}^{n_m}) = \frac{\mathbb{E}(\boldsymbol{S}\boldsymbol{A})-\mathbb{E}(\boldsymbol{S})\mathbb{E}(\boldsymbol{A})}{\sigma_{\boldsymbol{S}}\sigma_{\boldsymbol{A}}},
\label{eq:corr}
\end{align}
where $\sigma$ is the standard deviation.

\textbf{The deviation of Best Model}
\begin{align}
dev(\{\boldsymbol{S}_l\}_{l=1}^{n_m}, \{\boldsymbol{A}_l\}_{l=1}^{n_m}) = \max_l \boldsymbol{A}_l-\boldsymbol{A}_{l^*},
\label{eq:dev}
\end{align}
where $l^*=\operatorname{argmax}_l \boldsymbol{S}_l$ denotes the best model according to the metric. The metric with a higher correlation and lower deviation is more consistent with target accuracy.

\subsection{Derivation of Our Metrics}


\begin{figure}[t]
\vspace{-6mm}
\begin{minipage}[b]{0.38\textwidth}
    \centering
    \resizebox{0.9\textwidth}{!}{   
    \begin{tabular}{c|c|c}
        \hline
        Metric & CDAN & MCC \\
        \hline
        Source Acc. & 10.7 & 5.2 \\
        Entropy & 3.9 & 26.2 \\
        MI & 4.4 & 14.4 \\
        MI w. source & \textbf{0.3} & \textbf{0.5} \\
        \hline
    \end{tabular}}
    \captionof{table}{Using CDAN~\cite{CDAN} or MCC~\cite{MCC} as the training method, when search trade-off from \{0.1, 0.2, 0.3, 0.5, 1.0, 2.0, 3.0, 5.0, 10.0\}, ``dev'' of metrics. The results are averaged across 12 transfers in OfficeHome.}
    \label{tab:conclusion-source}
\end{minipage}
\hspace{0.02\textwidth}
\begin{minipage}[b]{0.6\textwidth}
	\centering
	\subfloat[CDAN]{
		\includegraphics[width=0.5\textwidth]{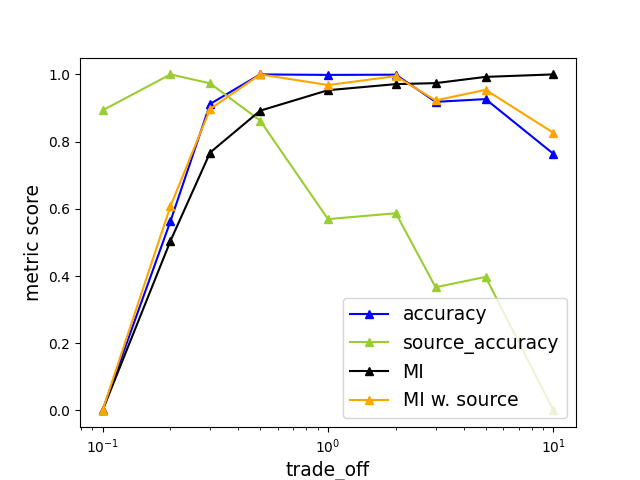}
    }
	\subfloat[MCC]{
		\includegraphics[width=0.5\textwidth]{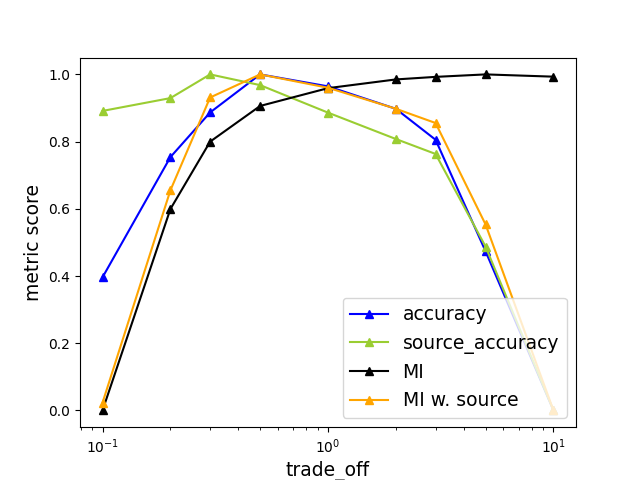}
	}
    \caption{On OfficeHome, using CDAN~\cite{CDAN} or MCC~\cite{MCC} as the training method, when the trade-off changes, the curves of various metrics. For display convenience, we normalize each metric to [0,1].}
	\label{fig:conclusion-source}
\end{minipage}
\vspace{-6mm}
\end{figure}

\subsubsection{Combine with Source Accuracy}
\label{sec:3.3.1}

Originating from Semi-supervised learning~\cite{SSLEntropy}, the Entropy of the prediction is commonly used in UDA methods~\cite{ADVENT} as a regularizer for unlabeled samples. Entropy can also serve as the evaluation metric for UDA~\cite{C-Ent}. As the Entropy metric is unaware of the ``mode collapse'' phenomenon, Mutual Information~\cite{MI} adds the diversity term. The Mutual Information metric (MI) is defined as: 
$\boldsymbol{MI} = H(\mathbb{E}_{\tilde{\boldsymbol{x}}^t}[\boldsymbol{p}^t]) - \mathbb{E}_{\tilde{\boldsymbol{x}}^t}[H(\boldsymbol{p}^t)]$, where $\boldsymbol{p}^t$ denotes the prediction of the model on $\tilde{\boldsymbol{x}}^t$. MI only considers the quality of the target samples. While 
the source label information and the quality of the source feature are ignored. In some situations, the prediction of the target sample is not aligned with our desired label space. To avoid this problem, SND~\cite{SND} suggests monitoring the source supervising loss or setting a threshold for source accuracy. However, we find this rough approach cannot help to determine the best model. To show the importance of source accuracy during evaluating UDA, we use CDAN~\cite{CDAN} or MCC~\cite{MCC} method to train the models with multiple trade-off hyper-parameters. We use the metric to evaluate the trained models and determine the best trade-off. As shown in Tab.~\ref{tab:conclusion-source} and Fig.~\ref{fig:conclusion-source}, the Entropy metric and MI increase as the trade-off increases, but the target accuracy first increases and then decreases as the trade-off increases. 
We propose to directly combine MI and source accuracy. We first normalize MI into $[0,1]$, then add it with the source accuracy, as follows:
\begin{align}
\boldsymbol{MI}_{w. source} = \mathbb{E}_{(\tilde{\boldsymbol{x}}^s,\tilde y^s)} I[\underset{k}{\operatorname{argmax}}
[\boldsymbol{p}^s] = \tilde y^s]+\frac{\boldsymbol{MI}}{2\log K}+\frac{1}{2},
\end{align}
where $\boldsymbol{p}^s$ denotes the prediction of the model on $\tilde{\boldsymbol{x}}^s$, and $K$ is the number of classes. 
As shown in Tab.~\ref{tab:conclusion-source} and Fig.~\ref{fig:conclusion-source}, this simple combination can balance MI on the target domain and source accuracy. If we view the MI term as the similarity between two domains, this metric formally follows Ben David's theory~\cite{BenDavidA}, where the source error and the domain discrepancy bound the target error.

\subsubsection{Robustness Property}
\label{sec:3.3.2}

\begin{figure}[t]
\vspace{-6mm}
	\centering
	\includegraphics[width=0.93\textwidth]{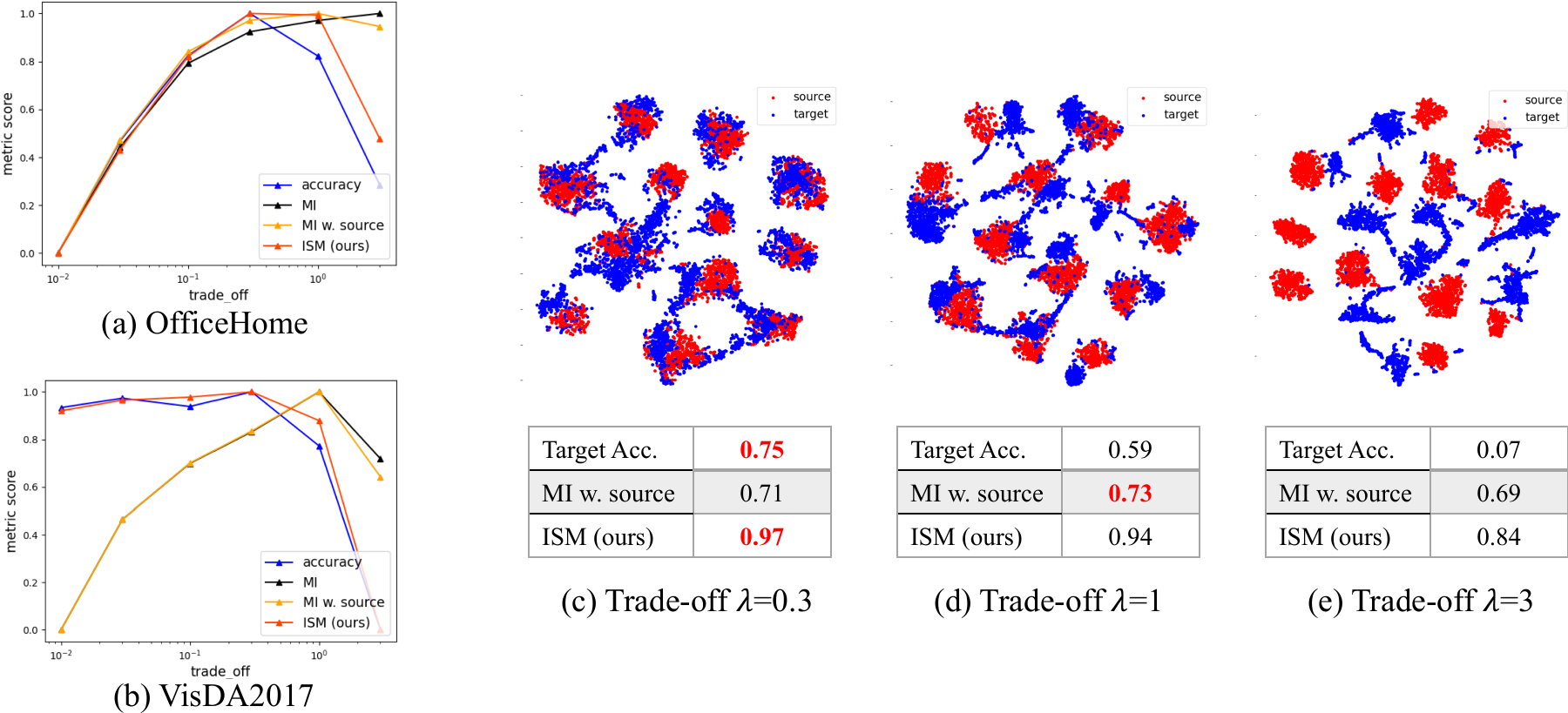}
	
    \caption{On OfficeHome and VisDA2017, use Mutual Information as the training algorithm. (a)-(b) When the trade-off value changes, the Mutual Information, ISM metric, and target domain accuracy change. (c)-(e) The tSNE~\cite{tSNE} visualization of model features and each metric scores when the trade-offs are equal to 0.3, 1, and 3 on VisDA2017.}
	\label{fig:conclusion-MI}
\vspace{-4mm}
\end{figure}

As mentioned in Section~\ref{sec:3.2}, a robust metric should not be vulnerable when we maliciously increase the metric score. To attack $\boldsymbol{MI}_{w. source}$ metric, we train the model with the following loss:
\begin{align}
 Loss = \mathbb{E}_{(\boldsymbol{x}^s,y^s)}[-\log \boldsymbol{p}_{y^s}]+\lambda (\sum_k \hat{\boldsymbol{p}}_k \log \hat{\boldsymbol{p}}_k-\mathbb{E}_{\boldsymbol{x}^t}[\sum_k \boldsymbol{p}_k \log \boldsymbol{p}_k]),
\end{align}
where $\hat{\boldsymbol{p}}_k$ is the average prediction for class $k$ within a batch. This loss is actually the UDA method used in ~\cite{MI}, which maximizes mutual information. We use this loss to train the models with different trade-off hyper-parameters $\lambda$.
As shown in Fig.~\ref{fig:conclusion-MI}, MI and MI w.source would prefer a large trade-off, but target accuracy decreases dramatically as the trade-off becomes larger. To investigate the cause, we use tSNE~\cite{tSNE} to visualize the source and target features. As demonstrated in Fig.~\ref{fig:conclusion-MI} (c)-(e), source features (red) form clear clusters, but target features are pushed away as the trade-off increases. The MI metric is almost unaware of this phenomenon because the training loss is finding the preference of MI.

The MI metric is vulnerable for two reasons: the evaluation metric is fully exposed to the training process, and the linear classifier $\textbf{f}$ cannot detect feature outliers. To solve this problem, we propose to train a new two-layer MLP on top of the source evaluation feature. Then we use the Mutual Information of this MLP classifier on the target evaluation feature as the evaluation metric. The new metric $\boldsymbol{ISM}$ can be formalized as follows:
\begin{align}
    \boldsymbol{IS}
	&= H(\mathbb{E}_{\tilde{\boldsymbol{x}}^t}[\boldsymbol{q}^t]) - \mathbb{E}_{\tilde{\boldsymbol{x}}^t}[H(\boldsymbol{q}^t)], \\
\boldsymbol{ISM} &= \mathbb{E}_{(\tilde{\boldsymbol{x}}^s,\tilde y^s)} I[\underset{k}{\operatorname{argmax}}
[\boldsymbol{p}^s] = \tilde y^s]+\frac{\boldsymbol{IS}}{2\log K}+\frac{1}{2},
\end{align}
where $\boldsymbol{q}^t=\textbf{h}(\textbf{g}(\tilde{\boldsymbol{x}}^t))$ and $\textbf{h}$ is the MLP classifier trained on the source evaluation feature. Noticeably, we still use the classifier of the model to compute the source accuracy, which evaluates whether the classifier $\textbf{f}$ is well-trained. As $\textbf{h}$ is held out during the UDA training and the two-layer MLP is more expressive, $\boldsymbol{ISM}$ is not vulnerable to attack. As shown in Fig.~\ref{fig:conclusion-MI}, when trained with the mutual information loss, $\boldsymbol{ISM}$ can be well consistent with the target accuracy. $\boldsymbol{ISM}$ is short for the Inception Score Metric for UDA because it is formally similar to the Inception Score for generative models~\cite{InceptionScore}. The inception score for the generative model utilizes the mutual information of the ImageNet pretrained inception network ~\cite{Inceptionv3}. Instead, we train an MLP classifier based on the supervision of the source evaluation set and combine it with the original source accuracy.

\subsubsection{Input-level Consistency}
\label{sec:3.3.3}

\begin{figure}[t]
\vspace{-6mm}
	\centering
	\subfloat[Train epochs of CDAN]{
		\includegraphics[width=0.25\textwidth]{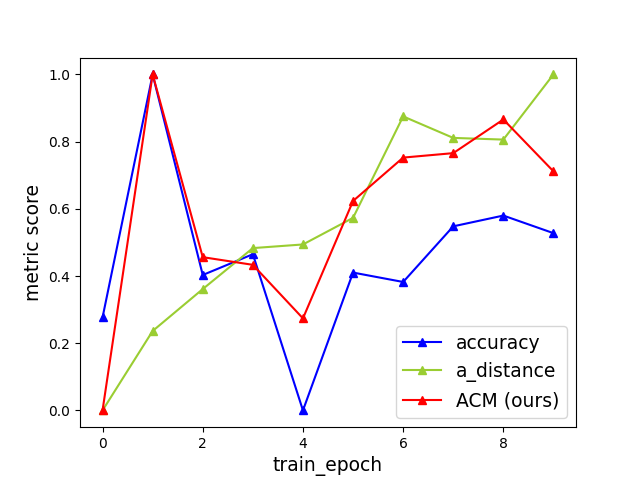}
        }
        \hspace{0.02\textwidth}
	\subfloat[Epoch 1]{
		\includegraphics[width=0.20\textwidth]{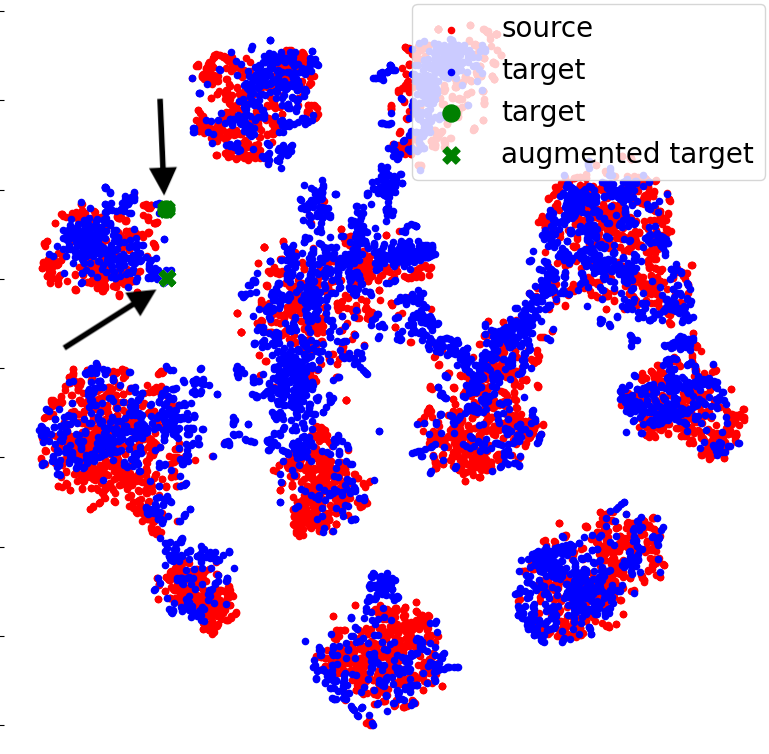}
	}
        \hspace{0.02\textwidth}
        \subfloat[Epoch 4]{
		\includegraphics[width=0.20\textwidth]{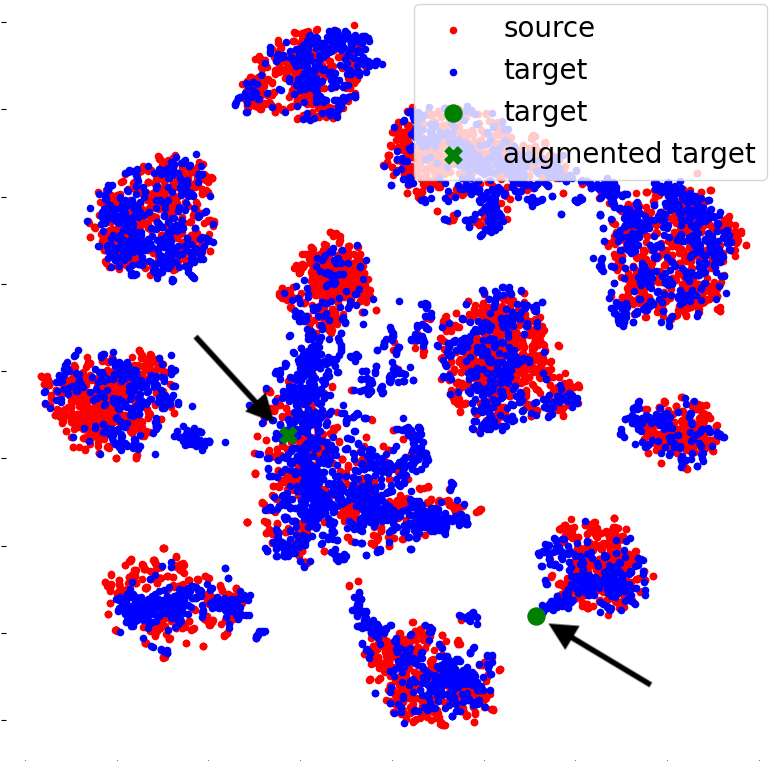}
        }
        \hspace{0.02\textwidth}
	\subfloat[Epoch 9]{
		\includegraphics[width=0.20\textwidth]{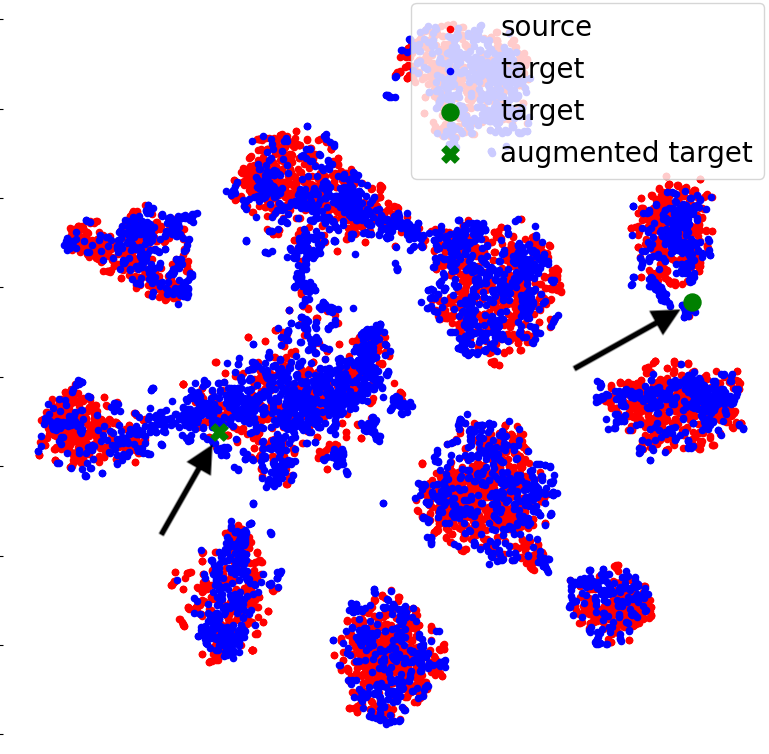}
	}
    \caption{On VisDA2017, using CDAN as the training method. (a) As training epochs grow, the curves of target accuracy, A-distance Metric, and ACM (ours). For display convenience, we normalize each metric to [0,1]. (b)-(d) The tSNE visualization of model features and a target feature (green circle) that gradually misclassified paired with the feature of its augmented view (green cross).}
\label{fig:conclusion-input}
\vspace{-4mm}
\end{figure}

In the experiments, our ISM already shows surprising consistency with target accuracy. However, we find that in some situations where we use feature alignment-based UDA methods, e.g., DANN and CDAN, the target accuracy might decrease with the training process. In these situations, the source and target features are well aligned, and source accuracy does not degrade, but target accuracy is not improving. We interpret this phenomenon as some target features being pulled to the wrong source class to make the target feature distribution the same as the source. We show this phenomenon in Fig~\ref{fig:conclusion-input}: the target accuracy approaches the maximum in the first epoch while the feature distributions of the two domains are not aligned. This phenomenon violates the common UDA assumption that samples embedded nearby are likely to share the labels~\cite{SND}. In these situations, the metrics that consider the features and predictions of two domains are hard to determine the model.

To solve this problem, we propose to use the consistency between the target sample and its augmented view to detect whether target features are over-aligned. This is based on the finding that for misaligned target samples, the features are more unstable to data augmentation. In Fig~\ref{fig:conclusion-input}, a target feature (green dot) is close to its data-augmented feature (green cross) at epoch 1, but the two become farther away as it is over-aligned. We define our Augment Consistency Metric (ACM) as follows:
\begin{align}
    \boldsymbol{AC}
	&= \mathbb{E}_{\tilde{\boldsymbol{x}}^t} I[\underset{k}{\operatorname{argmax}}[\boldsymbol{q}^t]= \underset{k}{\operatorname{argmax}}[\boldsymbol{q}^{t \prime}]] \\
	\boldsymbol{ACM} &= \mathbb{E}_{(\tilde{\boldsymbol{x}}^s,\tilde y^s)} I[\underset{k}{\operatorname{argmax}}[\boldsymbol{p}^s] = \tilde y^s]+\frac{1}{2}(\boldsymbol{AC} + \frac{H(\mathbb{E}_{\tilde{\boldsymbol{x}}^t}[\boldsymbol{q}^t])}{\log K}),
\end{align}
where $\boldsymbol{q}^{t \prime}=\textbf{h}(\textbf{g}(\tilde{\boldsymbol{x}}^{t \prime}))$ denotes the prediction of the MLP classifier on the data-augmented sample $\tilde{\boldsymbol{x}}^{t \prime}$. We use the MLP prediction instead of the original classifier to make it robust and combine it with the diversity term to avoid ``mode collapse''. As shown in Fig~\ref{fig:conclusion-input}, after taking input-level disturbance into consideration, ACM is consistent with target accuracy in the over-alignment situation. While input-level consistency has been utilized as a UDA method ~\cite{Selfensembling}, we are the first to study it as an evaluation metric.

\subsection{Flaws in Previous Metrics}
\label{sec:3.4}
In this section, we reveal the flaws in the experiment settings and metrics of previous works~\cite{DEV, SND}. To verify a robust evaluation metric for UDA, experimenting with sufficient datasets, training methods, and hyper-parameter sets are important. Based on the findings, we will construct our experiment settings in the experiment section~\ref{sec:4.1}.

\textbf{More Datasets and Training Methods are Important.}
In the DEV~\cite{DEV} and SND~\cite{SND} papers, they only used part of UDA datasets and part of UDA training methods, such as in the DEV paper ~\cite{DEV}, only the CDAN training method is used on Office31, and only the MCD training method is used on VisDA. In the SND paper ~\cite{SND}, although they used 4 training algorithms, they only test metrics on two transfer tasks of OfficeHome, Ar $\to$ Pr and Rw $\to$ Ar, and one transfer task of DomainNet, real $\to$ clipart.
However, \textbf{it is easy to draw wrong conclusions by testing evaluation metrics on the part of datasets.} 
As shown in the table ~\ref{tab:conclusion-dataset}, although DEV and SND perform well on some transfer tasks, e.g., Ar $\to$ Pr and Rw $\to$ Ar, they perform poorly on most transfer tasks. This is likely because the Ar $\to$ Pr and Rw $\to$ Ar transfer tasks are closer to the assumptions of their metric, e.g., SND assumes that tighter intra-class domains have higher accuracy. Our evaluation metric ACM achieves excellent results on all 12 transfer tasks. 

It is also important to validate the evaluation metrics using multiple training methods on each dataset. 
In the experiment, we employ five classic UDA algorithms. 
Additionally, we will investigate an interesting question: Can the metrics be used to select training methods for a transfer scenario? This problem is very important in practice because a large number of UDA algorithms have been proposed, so for a transfer scenario, it is troublesome to choose the most suitable training method. 


\begin{figure}[t]
\vspace{-6mm}
	\centering
	\subfloat[Learning Rate of MCC]{
		\includegraphics[width=0.24\textwidth]{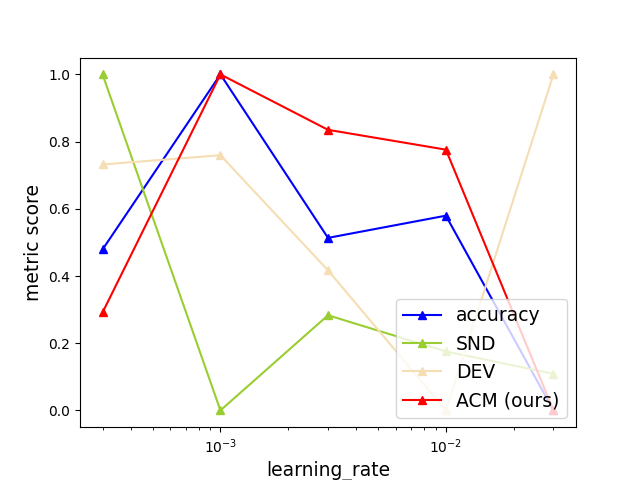}
        }
	\subfloat[Trade-off of MCC]{
		\includegraphics[width=0.24\textwidth]{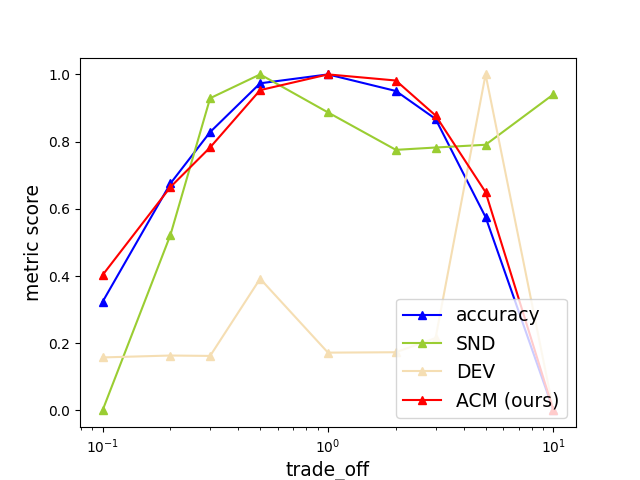}
	}
	\subfloat[Temperature of MCC]{
		\includegraphics[width=0.24\textwidth]{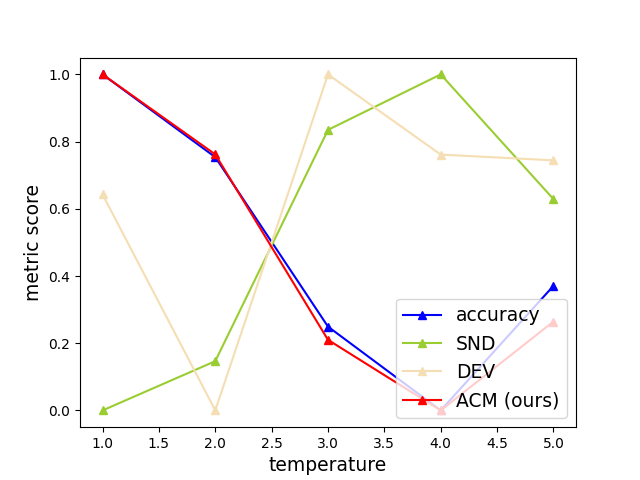}
	}
	\subfloat[Train Epoch of MCC]{
		\includegraphics[width=0.24\textwidth]{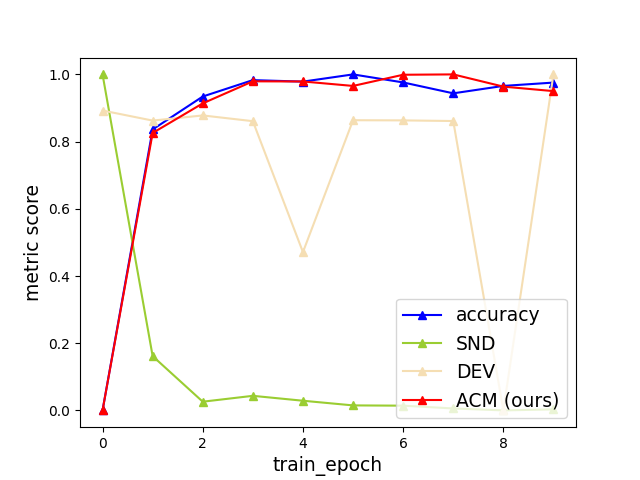}
	}
	\caption{On OfficeHome, when independently changing different hyper-parameters of MCC, curves of various evaluation metrics (averaged across 12 transfer tasks). We normalize each metric to [0,1]. }
	\label{fig:conclusion-hyperparameter}
\vspace{-2mm}
\end{figure}

\linespread{1.1}
\begin{table}[t]
    \centering
    
    \resizebox{1\textwidth}{!}{
    \begin{tabular}{|c|c|c|}
        \hline
        Training Method & Sparse hyper-parameter space & Dense hyper-parameter space \\
        \hline
        \multirow{2}{*}{Source only} & \{lr=\{1e-3,1e-2\}, wd=\{1e-4,1e-3\}, & \multirow{2}{*}{-} \\
        & train-epoch=\{1...10\}\} & \\
        \hline
        \multirow{3}{*}{DANN or CDAN} & 
        \multirow{3}{*}{\shortstack{\{lr=\{1e-3,1e-2\}, wd=\{1e-4,1e-3\}, \\ lr-multi-D=\{0.1 ,1,10\}, trade-off=\{0.1,1,10\} \\
        bottleneck-dim= \{256,512\}, train-epoch=\{1...10\} \}}} & 
        \multirow{3}{*}{\shortstack{\{lr=\{3e-4,1e-3,3e-3,1e-2,3e-2\}, wd=\{1e-4,3e- 4,1e-3\}, \\
        lr-multi-D=\{0.1,0.3,1,3,10\}, trade-off=\{0.1,0.3,1,3,10\}, \\
        bottleneck-dim= \{256,512\}, train-epoch=\{1...10\}\}}} \\
        & & \\
        & & \\
        \hline
        \multirow{3}{*}{MCC} & 
        \multirow{3}{*}{\shortstack{\{lr=\{1e-3,1e-2\}, wd=\{1e-4,1e-3\}, \\ trade-off=\{0.1,1,10\}, train-epoch=\{1...10\}\}}} & 
        \multirow{3}{*}{\shortstack{\{lr=\{3e-4,1e-3, 3e-3,1e-2,3e-2\}, wd=\{1e-4,3e-4,1e-3 \}, \\ temperature=\{1,2,3,4,5\}, trade-off=\{0.1,0.3,1,3,10\}, \\ bottleneck-dim=\{512,1024,2048 \}, train-epoch=\{1...10\}\}}} \\
        & & \\
        & & \\
        \hline
        \multirow{3}{*}{MDD} & 
        \multirow{3}{*}{\shortstack{\{lr=\{4e-4,4e-3\}, wd=\{5e-5,5e-4\}, \\ trade-off=\{0.1,1,10 \}, train-epoch=\{1...10\}\}}} & 
        \multirow{3}{*}{\shortstack{\{lr=\{3e-4,1e-3,3e-3,1e-2,3e-2\}, wd=\{1e-4,3e-4,1e-3 \}, \\ margin=\{1,2,3,4,5\}, trade-off=\{0.1,0.3,1,3,10\}, \\ bottleneck-dim=\{512,1024,2048 \}, train-epoch=\{1...10\}\}}}  \\
        & & \\
        & & \\
        \hline
    \end{tabular}}
    \linespread{1}
    \caption{The hyper-parameter spaces used in the experiments. 
    In the sparse hyper-parameter space, we perform grid search over all combinations of hyper-parameters to analyze the consistency. In the dense hyper-parameter space, we use our metric to search for optimal hyper-parameters.
    }
    \label{tab:hyperparameter}
\vspace{-6mm}
\end{table}
\linespread{1}

\textbf{Larger and Wider Hyper-parameter Sets are Important.}
For different datasets, the optimal hyper-parameters may vary by ten times~\cite{tllib}. In addition, it is often necessary to tune multiple hyper-parameters of the method to achieve the optimal result. However, in the previous DEV and SND papers, only one hyper-parameter was adjusted for each training method, and the adjustment range of hyper-parameters was small.
As shown in Fig.~\ref{fig:conclusion-hyperparameter}, when the hyper-parameter of MCC changes, previous metrics DEV and SND cannot be consistent with target accuracy. In addition, we also found that a larger selection interval will lead to different conclusions from the small ones. For example, when the trade-off values are changed, SND can maintain the same accuracy rate between 0.1 and 1.0, but if the trade-off value increases to 10.0, SND will give the opposite result. Finally, it can be seen from the figure that our ACM always maintains high consistency with target accuracy and can select the optimal value for various hyper-parameters.


\section{Experiments}


\subsection{Experimental Settings}
\label{sec:4.1}

\textbf{Datasets.} 
UDA datasets studied in the main paper: 
1) \textit{OfficeHome}~\cite{officehome} consists of 15,500 images with 65 classes from four domains: Artistic images (Ar), Clip art (Cl), Product images (Pr), and Real-world (Rw). There are 12 transfer tasks among these domains. 
2) \textit{VisDA2017}~\cite{visda} contains 12 categories and over 280,000 images from the Synthetic source domain and Real-world target domain.  
3) \textit{DomainNet}~\cite{domainnet} is a large-scale dataset for domain adaptation and contains 345 categories from six domains. We select four domains for our experiments: Clipart (c), Painting (p), Real (r), and Sketch (s). We only study single-source domain adaptation of DomainNet. There are 12 transfer tasks among these domains. 
4) \textit{Office31}~\cite{office} is a relatively small dataset containing 4652 images with 31 categories from three domains. Results on Office31 are provided in the Appendix. 

\textbf{Training Methods.} 
We use five popular UDA methods to get trained models. \textbf{1) Source only} \textbf{2) DANN}~\cite{DANN} \textbf{3) CDAN}~\cite{CDAN} \textbf{4) MDD}~\cite{MDD} \textbf{5) MCC}~\cite{MCC}.
The implementations of these methods all follow TL-Lib~\cite{tllib}.
For more implementation details, please refer to the Appendix.

\textbf{Sets of Hyper-parameters.} 
\label{sec:4.1.3}
We find that several hyper-parameters are often manually tuned, and we chose them to check the robustness of metrics. Totally we will change at most six hyper-parameters of the training method: \textbf{1) Early-stopping step} (train-epoch): For the UDA problem, the model at the final step is usually not the best model during training. 
We divide the total training step into ten epochs and evaluate the model after each epoch. 
\textbf{2) Learning rate} (lr): The initial learning rate \textbf{3) Weight decay} (wd) \textbf{4) Trade-off}: The trade-off between the supervised loss on the source domain and the target loss from UDA methods. \textbf{5) Bottleneck dimension}: The feature dimension output by the feature generator. \textbf{6) Hyper-parameter related to training methods}: We choose the margin $\gamma$~\cite{MDD} for MDD and the temperature $T$~\cite{MCC} for MCC. For DANN and CDAN, we tune the learning rate of the domain discriminator as the hyper-parameter to balance the convergence of the discriminator and the generator~\cite{FID}. We define lr-multi-D as the ratio of the learning rate of the discriminator to the generator.


\textbf{Unsupervised Evaluation Metrics.} $\mathcal{A}$-distance\cite{BenDavidA}, $\mathcal{H} \Delta \mathcal{H}$-divergence or MCD~\cite{BenDavidH,MCD}, MDD~\cite{MDD}, DEV~\cite{DEV}, Entropy~\cite{SSLEntropy, ADVENT}, SND~\cite{SND}, Mutual Information~\cite{MI}, ISM (ours), ACM (ours). We implement metrics according to the original papers and modify them to be positively correlated with target accuracy. The implementations are listed in the Appendix.

\begin{table*}[t]
\vspace{-6mm}
    \centering
    \resizebox{1\textwidth}{!}{
    \begin{tabular}{c|a|a|a|a|a|a}
        \toprule
        Training Method & \multicolumn{2}{|c}{Source only} & \multicolumn{2}{|c}{DANN} & \multicolumn{2}{|c}{CDAN} & \multicolumn{2}{|c}{MDD} & \multicolumn{2}{|c}{MCC} & \multicolumn{2}{|c}{ALL} \\
        \hline
        Metric & corr & dev & corr & dev & corr & dev & corr & dev & corr & dev & corr & dev \\
        \hline
        $\mathcal{A}$-distance & -0.81 & 6.99 & 0.55 & 6.47 & -0.17 & 6.56 & 0.58 & 2.25 & 0.37 & \textcolor{blue}{1.66} & 0.44 & 7.76 \\
        MCD & -0.58 & 8.03 & \textcolor{blue}{0.77} & 6.29 & -0.26 & 4.73 & \textcolor{red}{\textbf{0.86}} & 0.57 & -0.06 & 66.29 & 0.5 & 8.67 \\
        DEV & 0.12 & 7.39 & -0.08 & 4.37 & -0.08 & 4.71 & -0.09 & 47.54 & -0.11 & 64.27 & -0.03 & 64.27 \\
        Entropy & -0.14 & 8.99 & 0.56 & 4.81 & -0.28 & 9.34 & 0.64 & 1.17 & -0.06 & 66.29 & -0.34 & 66.29 \\
        SND & -0.74 & 8.12 & 0.46 & 8.51 & -0.72 & 9.81 & -0.55 & 50.33 & -0.58 & 67.65 & -0.42 & 52.12 \\
        MI & 0.06 & 6.26 & 0.58 & \textcolor{blue}{3.92} & -0.07 & 3.72 & 0.81 & \textcolor{red}{\textbf{0.0}} & 0.03 & 5.4 & 0.45 & 5.4 \\
        \hline
        ISM & \textcolor{red}{\textbf{0.84}} & \textcolor{red}{\textbf{0.31}} & 0.75 & \textcolor{blue}{3.92} & \textcolor{blue}{0.42} & \textcolor{blue}{1.23} & 0.75 & \textcolor{blue}{0.40} & \textcolor{blue}{0.88} & \textcolor{red}{\textbf{0.66}} & \textcolor{blue}{0.59} & \textcolor{red}{\textbf{1.66}} \\
        ACM & \textcolor{blue}{0.80} & \textcolor{blue}{2.38} & \textcolor{red}{\textbf{0.79}} & \textcolor{red}{\textbf{1.18}} & \textcolor{red}{\textbf{0.61}} & \textcolor{red}{\textbf{0.98}} & \textcolor{blue}{0.85} & \textcolor{red}{\textbf{0.0}} & \textcolor{red}{\textbf{0.93}} & \textcolor{blue}{1.66} & \textcolor{red}{\textbf{0.76}} & \textcolor{red}{\textbf{1.66}} \\
        \bottomrule
    \end{tabular}}
    \caption{Consistency between metrics of UDA and target accuracy on VisDA2017, when models are trained by different UDA methods and hyper-parameters.  The ``ALL'' method denotes assembling models trained by all five methods. The higher the Pearson's correlation (``corr'') and the lower the deviation (``dev''), the better the metric. \textcolor{red}{\textbf{Red score}} is the best and \textcolor{blue}{blue score} is the second best.}
    \label{tab:visda}
\vspace{-4mm}
\end{table*}

\begin{table*}[t]
    \centering
    \resizebox{1\textwidth}{!}{
    \begin{tabular}{c|a|a|a|a|a|a}
        \toprule
        Training Method & \multicolumn{2}{|c}{Source only} & \multicolumn{2}{|c}{DANN} & \multicolumn{2}{|c}{CDAN} & \multicolumn{2}{|c}{MDD} & \multicolumn{2}{|c}{MCC} & \multicolumn{2}{|c}{ALL} \\
        \hline
        Metric & corr & dev & corr & dev & corr & dev & corr & dev & corr & dev & corr & dev \\
        \hline
        $\mathcal{A}$-distance & 0.32 & 5.8 & 0.71 & \textcolor{blue}{1.5} & 0.67 & 1.82 & 0.93 & 1.27 & 0.45 & 8.62 & 0.56 & 6.71 \\
        MCD & 0.57 & \textcolor{red}{\textbf{1.12}} & \textcolor{blue}{0.75} & 2.55 & 0.69 & 1.79 & 0.93 & \textcolor{blue}{1.13} & \textcolor{blue}{0.76} & \textcolor{red}{\textbf{1.03}} & 0.71 & 3.16 \\
        DEV & 0.01 & 4.32 & 0.06 & 8.14 & 0.06 & 3.4 & 0.11 & 9.54 & -0.02 & 11.51 & 0.01 & 12.42 \\
        Entropy & -0.64 & 8.20 & 0.43 & 4.28 & 0.88 & 1.56 & 0.88 & 1.98 & 0.38 & 24.43 & 0.52 & 24.52 \\
        SND & -0.60 & 8.17 & -0.23 & 7.90 & 0.07 & 9.24 & -0.90 & 54.57 & -0.20 & 12.40 & -0.25 & 34.75 \\
        MI & -0.60 & 6.78 & 0.45 & 4.25 & 0.88 & 1.40 & 0.91 & 1.98 & 0.37 & 22.83 & 0.52 & 22.93 \\
        \hline
        ISM & \textcolor{blue}{0.72} & 1.47 & 0.6 & 1.68 & \textcolor{red}{\textbf{0.91}} & \textcolor{blue}{1.15} & \textcolor{red}{\textbf{0.97}} & 1.45 & 0.70 & 1.96 & \textcolor{blue}{0.88} & \textcolor{blue}{1.96} \\
        ACM & \textcolor{red}{\textbf{0.75}} & \textcolor{blue}{1.37} & \textcolor{red}{\textbf{0.77}} & \textcolor{red}{\textbf{1.16}} & \textcolor{blue}{0.90} & \textcolor{red}{\textbf{1.13}} & \textcolor{blue}{0.95} & \textcolor{red}{\textbf{0.93}} & \textcolor{red}{\textbf{0.94}} & \textcolor{blue}{1.36} & \textcolor{red}{\textbf{0.93}} & \textcolor{red}{\textbf{1.73}} \\
        \bottomrule
    \end{tabular}}
    \caption{The ``corr'' and ``dev'' results are averaged over the 12 transfer tasks of OfficeHome. }
    \label{tab:officehome}
\vspace{-4mm}
\end{table*}

\subsection{Main Results}
\label{sec:4.2}

In this section, we investigate whether unsupervised evaluation metrics satisfy the ``Consistency'' principle in Section ~\ref{sec:3.2}. We train the model $\{\boldsymbol{M}_l\}_{l=1}^{n_m}$ using the five UDA methods and the hyper-parameters for the coarse hyper-parameter space in the Tab.~\ref{tab:hyperparameter}. For each metric, we report the Pearson correlation (``corr'') and the deviation of the best model (``dev'').
Tab. ~\ref{tab:visda}, Tab. ~\ref{tab:domainnet}, and Tab. ~\ref{tab:officehome} show the results of UDA metrics for five training methods on VisDA2017, OfficeHome, and DomainNet. As the results show, it is difficult for previous metrics to represent the target accuracy across all training methods. Some metrics can perform well on the transfer task on one of the datasets but did not perform well on all three, which also shows that testing on partial datasets may lead to biased conclusions. 
Notably, our proposed ISM is consistent with the target accuracy for most training methods. Our ACM achieves better performance for training methods that align features of two domains, e.g., DANN and CDAN, as it can detect the over-alignment problem.

\textbf{Comparison of training methods:}
We also investigate the consistency of metrics when comparing different methods. Because in practice, we need to determine the best UDA method for the transfer task. We collected all models trained by all five methods with their metric scores and target accuracy. For each metric, we compute Pearson's correlation and the deviation of the best model, and the results are shown in the "ALL" column.
As shown in Tables ~\ref{tab:visda}, Table ~\ref{tab:domainnet}, and Table ~\ref{tab:officehome}, when comparing all training methods, maintaining consistency has become more difficult for most metrics. It is worth noting that our ISM and ACM perform well on all three datasets, with the deviation of the best model (``dev'') below 2\%. Therefore, we can use the proposed unsupervised metrics to decide the best training method and its hyper-parameters for a dataset.

\textbf{Robustness property} We show the robustness property of ISM and ACM in the Appendix.

\begin{table*}[t]
\vspace{-6mm}
    \centering
    \resizebox{1\textwidth}{!}{
    \begin{tabular}{c|a|a|a|a|a|a}
        \toprule
        Training Method & \multicolumn{2}{|c}{Source only} & \multicolumn{2}{|c}{DANN} & \multicolumn{2}{|c}{CDAN} & \multicolumn{2}{|c}{MDD} & \multicolumn{2}{|c}{MCC} & \multicolumn{2}{|c}{ALL} \\
        \hline
        Metric & corr & dev & corr & dev & corr & dev & corr & dev & corr & dev & corr & dev \\
        \hline
        $\mathcal{A}$-distance & \textcolor{blue}{0.89} & 1.89 & 0.64 & 0.9 & 0.83 & \textcolor{blue}{0.38} & \textcolor{blue}{0.93} & 0.45 & 0.6 & 9.45 & \textcolor{blue}{0.89} & 4.19 \\
        MCD & 0.87 & 1.74 & 0.67 & 6.48 & 0.95 & \textcolor{blue}{0.38} & 0.89 & 0.45 & \textcolor{red}{\textbf{0.94}} & 5.05 & 0.86 & 13.57 \\
        DEV & 0.19 & 1.53 & 0.07 & 3.0 & 0.08 & 1.44 & -0.04 & 12.25 & -0.11 & 5.14 & 0.0 & 2.45 \\
        Entropy & 0.45 & 3.43 & 0.65 & 5.83 & 0.79 & 0.49 & 0.83 & 1.09 & 0.75 & 1.37 & 0.71 & 3.55 \\
        SND & -0.93 & 11.8 & -0.95 & 20.46 & -0.95 & 11.2 & -0.98 & 39.8 & -0.81 & 24.27 & -0.75 & 23.96 \\
        MI & 0.48 & 3.21 & 0.65 & 5.83 & 0.79 & 0.49 & 0.92 & 0.87 & 0.76 & 0.93 & 0.71 & 3.11 \\
        \hline
        ISM & 0.85 & \textcolor{red}{\textbf{1.33}} & \textcolor{red}{\textbf{0.87}} & \textcolor{blue}{0.7} & \textcolor{blue}{0.96} & 0.41 & \textcolor{red}{\textbf{0.98}} & \textcolor{blue}{0.28} & \textcolor{blue}{0.92} & \textcolor{blue}{0.6} & \textcolor{red}{\textbf{0.91}} & \textcolor{blue}{1.13} \\
        ACM & \textcolor{red}{\textbf{0.94}} & \textcolor{blue}{1.41} & \textcolor{blue}{0.8} & \textcolor{red}{\textbf{0.27}} & \textcolor{red}{\textbf{0.98}} & \textcolor{red}{\textbf{0.29}} & \textcolor{blue}{0.93} & \textcolor{red}{\textbf{0.04}} & 0.84 & \textcolor{red}{\textbf{0.15}} & 0.87 & \textcolor{red}{\textbf{0.29}} \\
        \bottomrule
    \end{tabular}}
    \caption{The ``corr'' and ``dev'' results are averaged over 12 transfer tasks of DomainNet. }
    \label{tab:domainnet}
\vspace{-4mm}
\end{table*}

\begin{table*}[t]
	\centering
	\resizebox{0.98\textwidth}{!}{
		\begin{tabular}{cccccccccccccc}  
			\toprule
		Method & plane & bcycl & bus & car & house &  knife & mcycl & person & plant & sktbrd & train & truck & Avg    \\
			\hline
			DANN (default) & 81.7 & 38.7 & 77.8 & 85.8 & 67.2 & 76.7 & 65.5 & 57.9 & 81.3 & 50.4 & 88.5 & 61.0 & 69.4 \\
			DANN (searched) & 84.8 & 45.5 & 86.9 & 86.8 & 74.0 & 91.3 & 75.7 & 59.7 & 89.9 & 51.2 & 82.3 & 62.2 & 74.2\\
			Gains ($+\Delta$) & \textcolor{red}{+3.1} & \textcolor{red}{+6.8} & \textcolor{red}{+9.1} & \textcolor{red}{+1.0} & \textcolor{red}{+6.8} & \textcolor{red}{+14.6} & \textcolor{red}{+10.2} & \textcolor{red}{+1.8} & \textcolor{red}{+8.6} & \textcolor{red}{+0.8} & \textcolor{green}{-6.2} & \textcolor{red}{+1.2} & \textcolor{red}{\textbf{+4.8}} \\
			\hline
			CDAN (default) & 84.8 & 51.5 & 78.8 & 85.1 & 70.0 & 90.8 & 69.7 & 58.6 & 88.2 & 48.5 & 80.2 & 65.3 & 72.6 \\
			CDAN (searched) & 86.6 & 47.0 & 82.6 & 85.9 & 75.6 & 87.0 & 78.0 & 63.5 & 88.2 & 55.0 & 79.6 & 76.2 & 75.4 \\
			Gains ($+\Delta$) & \textcolor{red}{+1.8} & \textcolor{green}{-4.5} & \textcolor{red}{+3.8} & \textcolor{red}{+0.8} & \textcolor{red}{+5.6} & \textcolor{green}{-3.8} & \textcolor{red}{+8.3} & \textcolor{red}{+4.9} & \textcolor{red}{+0.0} & \textcolor{red}{+6.5} & \textcolor{green}{-0.6} & \textcolor{red}{+10.9} & \textcolor{red}{\textbf{+2.8}} \\
			\hline
			MDD (default) & 68.9 & 59.5 & 89.7 & 89.5 & 67.8 & 94.4 & 73.7 & 50.2 & 93.4 & 59.0 & 79.5 & 66.2 & 74.3 \\
			MDD (searched) & 82.0 & 54.6 & 86.9 & 90.7 & 81.4 & 94.6 & 78.2 & 64.4 & 88.4 & 57.2 & 83.1 & 69.4 & 77.6 \\
			Gains ($+\Delta$) & \textcolor{red}{+13.1} & \textcolor{green}{-4.9} & \textcolor{green}{-2.8} & \textcolor{red}{+1.2} & \textcolor{red}{+13.6} & \textcolor{red}{+0.2} & \textcolor{red}{+4.5} & \textcolor{red}{+14.2} & \textcolor{green}{-5.0} & \textcolor{green}{-1.8} & \textcolor{red}{+3.6} & \textcolor{red}{+3.2} & \textcolor{red}{\textbf{+3.3}} \\
			\hline
			MCC (default) & 85.9 & 71.1 & 77.9 & 87.1 & 80.1 & 82.6 & 58.9 & 58.8 & 90.2 & 55.8 & 80.7 & 75.1 & 75.3 \\
			MCC (searched) & 88.5 & 69.3 & 79.2 & 91.0 & 81.7 & 85.0 & 71.0 & 64.3 & 92.8 & 61.1 & 80.0 & 77.2 & 78.4 \\
                Gains ($+\Delta$) & \textcolor{red}{+2.6} & \textcolor{green}{-1.8} & \textcolor{red}{+1.3} & \textcolor{red}{+3.9} & \textcolor{red}{+1.6} & \textcolor{red}{+2.4} & \textcolor{red}{+12.1} & \textcolor{red}{+5.5} & \textcolor{red}{+2.6} & \textcolor{red}{+5.3} & \textcolor{green}{-0.7} & \textcolor{red}{+2.1} & \textcolor{red}{\textbf{+3.1}} \\
			\bottomrule
	\end{tabular}}
        \caption{Hyper-parameters found by our metric vs. those manually tuned on VisDA2017. }
	\label{tab:visda-search}
\vspace{-4mm}
\end{table*}

\subsection{Unsupervised Hyper-parameter Search}

Most UDA methods require manual tuning of hyper-parameters for different datasets. It would be ideal to unsupervised find suitable hyper-parameters automatically. In this section, we show that our ACM can be used for the unsupervised search of hyper-parameters. We will conduct unsupervised hyper-parameter searches for four algorithms: DANN, CDAN, MCC, and MDD. For each UDA training method, we first define its hyper-parameter search space, shown in the dense hyper-parameter space in Tab.~\ref{tab:hyperparameter}. 
We set ACM as the target of the hyper-parameter search. We simply utilized the TPE search algorithm ~\cite{TPE} for 50 trials and Optuna's median pruner ~\cite{Optuna} to speed up the search. For each transfer task in the dataset, we report the target accuracy of the best model found by ACM. We compare this to the performance of the default hyper-parameters for each method used in the TL-Lib~\cite{tllib}.
Tab.~\ref{tab:visda-search} shows the target accuracy of the model found by our metric and the default model on VisDA. For all four training methods, the hyper-parameters found by us outperform those manually tuned by TL-Lib. Unlike previous supervised tuning, our search process requires no label information on the target domain. Results on Office and DomainNet can be found in the Appendix.


\section{Conclusion}

This paper studies the principles that a robust UDA evaluation metric satisfies. By analyzing the drawbacks of the mutual information metric, we propose Inception Score Metric for UDA (ISM) and Augmentation Consistency Metric (ACM). By conducting extensive experiments, we validate the effectiveness of our metrics in a variety of scenarios. Additionally, our research highlights the potential of evaluation metrics to further the development of AutoML in the UDA.

	\bibliographystyle{iclr2024}
	\bibliography{egbib}

\begin{thebibliography}{48}
\providecommand{\natexlab}[1]{#1}
\providecommand{\url}[1]{\texttt{#1}}
\expandafter\ifx\csname urlstyle\endcsname\relax
  \providecommand{\doi}[1]{doi: #1}\else
  \providecommand{\doi}{doi: \begingroup \urlstyle{rm}\Url}\fi

\bibitem[Akiba et~al.(2019)Akiba, Sano, Yanase, Ohta, and Koyama]{Optuna}
Takuya Akiba, Shotaro Sano, Toshihiko Yanase, Takeru Ohta, and Masanori Koyama.
\newblock Optuna: A next-generation hyperparameter optimization framework.
\newblock \emph{Proceedings of the 25th ACM SIGKDD International Conference on
  Knowledge Discovery \& Data Mining}, 2019.

\bibitem[Ben-David et~al.(2006)Ben-David, Blitzer, Crammer, and
  Pereira]{BenDavidA}
Shai Ben-David, John Blitzer, Koby Crammer, and Fernando~C Pereira.
\newblock Analysis of representations for domain adaptation.
\newblock In \emph{NeurIPS}, 2006.

\bibitem[Ben-David et~al.(2010)Ben-David, Blitzer, Crammer, Kulesza, Pereira,
  and Vaughan]{BenDavidH}
Shai Ben-David, John Blitzer, Koby Crammer, Alex Kulesza, Fernando~C Pereira,
  and Jennifer~Wortman Vaughan.
\newblock A theory of learning from different domains.
\newblock \emph{Machine Learning}, 79:\penalty0 151--175, 2010.

\bibitem[Bergstra et~al.(2011)Bergstra, Bardenet, Bengio, and K{\'e}gl]{TPE}
James Bergstra, R{\'e}mi Bardenet, Yoshua Bengio, and Bal{\'a}zs K{\'e}gl.
\newblock Algorithms for hyper-parameter optimization.
\newblock In \emph{NeurIPS}, 2011.

\bibitem[Carion et~al.(2020)Carion, Massa, Synnaeve, Usunier, Kirillov, and
  Zagoruyko]{DETR}
Nicolas Carion, Francisco Massa, Gabriel Synnaeve, Nicolas Usunier, Alexander
  Kirillov, and Sergey Zagoruyko.
\newblock End-to-end object detection with transformers.
\newblock \emph{ArXiv}, abs/2005.12872, 2020.
\newblock URL \url{https://api.semanticscholar.org/CorpusID:218889832}.

\bibitem[Chen et~al.(2018)Chen, Zhu, Papandreou, Schroff, and Adam]{Deeplabv3+}
Liang-Chieh Chen, Yukun Zhu, George Papandreou, Florian Schroff, and Hartwig
  Adam.
\newblock Encoder-decoder with atrous separable convolution for semantic image
  segmentation.
\newblock In \emph{ECCV}, 2018.

\bibitem[Cui et~al.(2020)Cui, Wang, Zhuo, Li, Huang, and Tian]{BNM}
Shuhao Cui, Shuhui Wang, Junbao Zhuo, Liang Li, Qingming Huang, and Qi~Tian.
\newblock Towards discriminability and diversity: Batch nuclear-norm
  maximization under label insufficient situations.
\newblock In \emph{CVPR}, 2020.

\bibitem[Dinu et~al.(2023)Dinu, Holzleitner, Beck, Nguyen, Huber, Eghbal-zadeh,
  Moser, Pereverzyev, Hochreiter, and Zellinger]{Aggregation}
Marius-Constantin Dinu, Markus Holzleitner, Maximilian~Heinz Beck, Hoan~Duc
  Nguyen, Andrea Huber, Hamid Eghbal-zadeh, Bernhard~Alois Moser, Sergei~V.
  Pereverzyev, Sepp Hochreiter, and Werner Zellinger.
\newblock Addressing parameter choice issues in unsupervised domain adaptation
  by aggregation.
\newblock \emph{ArXiv}, abs/2305.01281, 2023.

\bibitem[French et~al.(2017)French, Mackiewicz, and Fisher]{Selfensembling}
Geoffrey French, Michal Mackiewicz, and Mark~H. Fisher.
\newblock Self-ensembling for visual domain adaptation.
\newblock In \emph{ICLR}, 2017.

\bibitem[Ganin et~al.(2016)Ganin, Ustinova, Ajakan, Germain, Larochelle,
  Laviolette, Marchand, and Lempitsky]{DANN}
Yaroslav Ganin, E.~Ustinova, Hana Ajakan, Pascal Germain, H.~Larochelle,
  François Laviolette, Mario Marchand, and Victor~S. Lempitsky.
\newblock Domain-adversarial training of neural networks.
\newblock In \emph{Journal of Machine Learning Research}, 2016.

\bibitem[Goodfellow et~al.(2014{\natexlab{a}})Goodfellow, Shlens, and
  Szegedy]{AdversarialExamples}
Ian~J. Goodfellow, Jonathon Shlens, and Christian Szegedy.
\newblock Explaining and harnessing adversarial examples.
\newblock \emph{CoRR}, abs/1412.6572, 2014{\natexlab{a}}.

\bibitem[Goodfellow et~al.(2014{\natexlab{b}})Goodfellow, Shlens, and
  Szegedy]{Goodfellow2014ExplainingAH}
Ian~J. Goodfellow, Jonathon Shlens, and Christian Szegedy.
\newblock Explaining and harnessing adversarial examples.
\newblock \emph{CoRR}, abs/1412.6572, 2014{\natexlab{b}}.

\bibitem[Grandvalet \& Bengio(2004)Grandvalet and Bengio]{SSLEntropy}
Yves Grandvalet and Yoshua Bengio.
\newblock Semi-supervised learning by entropy minimization.
\newblock In \emph{NeurIPS}, 2004.

\bibitem[Gu et~al.(2021)Gu, Lin, Kuo, and Cui]{ViLD}
Xiuye Gu, Tsung-Yi Lin, Weicheng Kuo, and Yin Cui.
\newblock Open-vocabulary object detection via vision and language knowledge
  distillation.
\newblock In \emph{ICLR}, 2021.

\bibitem[He et~al.(2016)He, Zhang, Ren, and Sun]{ResNet}
Kaiming He, Xiangyu Zhang, Shaoqing Ren, and Jian Sun.
\newblock Deep residual learning for image recognition.
\newblock In \emph{CVPR}, 2016.

\bibitem[Heusel et~al.(2017)Heusel, Ramsauer, Unterthiner, Nessler, and
  Hochreiter]{FID}
Martin Heusel, Hubert Ramsauer, Thomas Unterthiner, Bernhard Nessler, and Sepp
  Hochreiter.
\newblock Gans trained by a two time-scale update rule converge to a local nash
  equilibrium.
\newblock In \emph{NeurIPS}, 2017.

\bibitem[Jiang et~al.(2020)Jiang, Chen, Fu, and Long]{tllib}
Junguang Jiang, Baixu Chen, Bo~Fu, and Mingsheng Long.
\newblock Transfer-learning-library.
\newblock \url{https://github.com/thuml/Transfer-Learning-Library}, 2020.

\bibitem[Jin et~al.(2020)Jin, Wang, Long, and Wang]{MCC}
Ying Jin, Ximei Wang, Mingsheng Long, and Jianmin Wang.
\newblock Minimum class confusion for versatile domain adaptation.
\newblock In \emph{ECCV}, 2020.

\bibitem[Kang et~al.(2019)Kang, Jiang, Yang, and Hauptmann]{CAN}
Guoliang Kang, Lu~Jiang, Yi~Yang, and Alexander Hauptmann.
\newblock Contrastive adaptation network for unsupervised domain adaptation.
\newblock In \emph{CVPR}, 2019.

\bibitem[Liu et~al.(2022)Liu, Mao, Wu, Feichtenhofer, Darrell, and
  Xie]{ConvNeXt}
Zhuang Liu, Hanzi Mao, Chaozheng Wu, Christoph Feichtenhofer, Trevor Darrell,
  and Saining Xie.
\newblock A convnet for the 2020s.
\newblock \emph{CVPR}, 2022.

\bibitem[Long et~al.(2015)Long, Cao, Wang, and Jordan]{DAN}
Mingsheng Long, Yue Cao, Jianmin Wang, and Michael~I. Jordan.
\newblock Learning transferable features with deep adaptation networks.
\newblock In \emph{ICML}, 2015.

\bibitem[Long et~al.(2018)Long, Cao, Wang, and Jordan]{CDAN}
Mingsheng Long, Zhangjie Cao, Jianmin Wang, and Michael~I. Jordan.
\newblock Conditional adversarial domain adaptation.
\newblock In \emph{NeurIPS}, 2018.

\bibitem[Morerio et~al.(2017)Morerio, Cavazza, and Murino]{C-Ent}
Pietro Morerio, Jacopo Cavazza, and Vittorio Murino.
\newblock Minimal-entropy correlation alignment for unsupervised deep domain
  adaptation.
\newblock \emph{ArXiv}, abs/1711.10288, 2017.

\bibitem[Musgrave et~al.(2022)Musgrave, Belongie, and Lim]{ThreeNV}
Kevin Musgrave, Serge~J. Belongie, and Ser~Nam Lim.
\newblock Three new validators and a large-scale benchmark ranking for
  unsupervised domain adaptation.
\newblock \emph{ArXiv}, 2022.

\bibitem[Peng et~al.(2017)Peng, Usman, Kaushik, Hoffman, Wang, and
  Saenko]{visda}
Xingchao Peng, Ben Usman, Neela Kaushik, Judy Hoffman, Dequan Wang, and Kate
  Saenko.
\newblock Visda: The visual domain adaptation challenge.
\newblock \emph{ArXiv}, abs/1710.06924, 2017.

\bibitem[Peng et~al.(2019)Peng, Bai, Xia, Huang, Saenko, and Wang]{domainnet}
Xingchao Peng, Qinxun Bai, Xide Xia, Zijun Huang, Kate Saenko, and Bo~Wang.
\newblock Moment matching for multi-source domain adaptation.
\newblock In \emph{ICCV}, 2019.

\bibitem[Pham et~al.(2018)Pham, Guan, Zoph, Le, and Dean]{ENAS}
Hieu Pham, Melody~Y. Guan, Barret Zoph, Quoc~V. Le, and Jeff Dean.
\newblock Efficient neural architecture search via parameter sharing.
\newblock In \emph{ICML}, 2018.

\bibitem[Radford et~al.(2021)Radford, Kim, Hallacy, Ramesh, Goh, Agarwal,
  Sastry, Askell, Mishkin, Clark, Krueger, and Sutskever]{CLIP}
Alec Radford, Jong~Wook Kim, Chris Hallacy, Aditya Ramesh, Gabriel Goh,
  Sandhini Agarwal, Girish Sastry, Amanda Askell, Pamela Mishkin, Jack Clark,
  Gretchen Krueger, and Ilya Sutskever.
\newblock Learning transferable visual models from natural language
  supervision.
\newblock In \emph{ICML}, 2021.

\bibitem[Saenko \& Kulis(2010)Saenko and Kulis]{office}
Kate Saenko and Brian Kulis.
\newblock Adapting visual category models to new domains.
\newblock In \emph{ECCV}, 2010.

\bibitem[Saito et~al.(2018)Saito, Watanabe, Ushiku, and Harada]{MCD}
Kuniaki Saito, Kohei Watanabe, Y.~Ushiku, and Tatsuya Harada.
\newblock Maximum classifier discrepancy for unsupervised domain adaptation.
\newblock In \emph{CVPR}, 2018.

\bibitem[Saito et~al.(2021)Saito, Kim, Teterwak, Sclaroff, Darrell, and
  Saenko]{SND}
Kuniaki Saito, Donghyun Kim, Piotr Teterwak, Stan Sclaroff, Trevor Darrell, and
  Kate Saenko.
\newblock Tune it the right way: Unsupervised validation of domain adaptation
  via soft neighborhood density.
\newblock In \emph{ICCV}, 2021.

\bibitem[Salimans et~al.(2016)Salimans, Goodfellow, Zaremba, Cheung, Radford,
  and Chen]{InceptionScore}
Tim Salimans, Ian~J. Goodfellow, Wojciech Zaremba, Vicki Cheung, Alec Radford,
  and Xi~Chen.
\newblock Improved techniques for training gans.
\newblock In \emph{NeurIPS}, 2016.

\bibitem[She et~al.(2020)She, Feng, Hao, Yang, Lan, Lomonaco, Shi, Wang, Guo,
  Zhang, Qiao, and Chan]{RobotLifelong}
Qi~She, Fan Feng, Xinyue Hao, Qihan Yang, Chuanlin Lan, Vincenzo Lomonaco,
  Xuesong Shi, Zhengwei Wang, Yao Guo, Yimin Zhang, Fei Qiao, and Rosa H.~M.
  Chan.
\newblock Openloris-object: A robotic vision dataset and benchmark for lifelong
  deep learning.
\newblock In \emph{ICRA}, 2020.

\bibitem[Shi \& Sha(2012)Shi and Sha]{MI}
Yuan Shi and Fei Sha.
\newblock Information-theoretical learning of discriminative clusters for
  unsupervised domain adaptation.
\newblock In \emph{ICML}, 2012.

\bibitem[Sugiyama et~al.(2007)Sugiyama, Krauledat, and M{\"u}ller]{IWV}
Masashi Sugiyama, Matthias Krauledat, and Klaus-Robert M{\"u}ller.
\newblock Covariate shift adaptation by importance weighted cross validation.
\newblock \emph{JMLR}, 8:\penalty0 985--1005, 2007.

\bibitem[Sun \& Saenko(2016)Sun and Saenko]{Deepcoral}
Baochen Sun and Kate Saenko.
\newblock Deep coral: Correlation alignment for deep domain adaptation.
\newblock In \emph{ECCV Workshops}, 2016.

\bibitem[Szegedy et~al.(2016)Szegedy, Vanhoucke, Ioffe, Shlens, and
  Wojna]{Inceptionv3}
Christian Szegedy, Vincent Vanhoucke, Sergey Ioffe, Jonathon Shlens, and
  Zbigniew Wojna.
\newblock Rethinking the inception architecture for computer vision.
\newblock In \emph{CVPR}, 2016.

\bibitem[Tanwisuth et~al.(2021)Tanwisuth, Fan, Zheng, Zhang, Zhang, Chen, and
  Zhou]{Proto}
Korawat Tanwisuth, Xinjie Fan, Huangjie Zheng, Shujian Zhang, Hao Zhang,
  Bo~Chen, and Mingyuan Zhou.
\newblock A prototype-oriented framework for unsupervised domain adaptation.
\newblock In \emph{NeurIPS}, 2021.

\bibitem[van~der Maaten \& Hinton(2008)van~der Maaten and Hinton]{tSNE}
Laurens van~der Maaten and Geoffrey~E. Hinton.
\newblock Visualizing data using t-sne.
\newblock In \emph{JMLR}, 2008.

\bibitem[Venkateswara et~al.(2017)Venkateswara, Eusebio, Chakraborty, and
  Panchanathan]{officehome}
Hemanth Venkateswara, Jose Eusebio, Shayok Chakraborty, and Sethuraman
  Panchanathan.
\newblock Deep hashing network for unsupervised domain adaptation.
\newblock In \emph{CVPR}, 2017.

\bibitem[Vu et~al.(2019)Vu, Jain, Bucher, Cord, and P{\'e}rez]{ADVENT}
Tuan-Hung Vu, Himalaya Jain, Max Bucher, Matthieu Cord, and Patrick P{\'e}rez.
\newblock Advent: Adversarial entropy minimization for domain adaptation in
  semantic segmentation.
\newblock In \emph{CVPR}, 2019.

\bibitem[Xie et~al.(2021)Xie, Wang, Yu, Anandkumar, {\'A}lvarez, and
  Luo]{SegFormer}
Enze Xie, Wenhai Wang, Zhiding Yu, Anima Anandkumar, Jos{\'e}~Manuel
  {\'A}lvarez, and Ping Luo.
\newblock Segformer: Simple and efficient design for semantic segmentation with
  transformers.
\newblock \emph{ArXiv}, abs/2105.15203, 2021.

\bibitem[You et~al.(2019)You, Wang, Long, and Jordan]{DEV}
Kaichao You, Ximei Wang, Mingsheng Long, and Michael~I. Jordan.
\newblock Towards accurate model selection in deep unsupervised domain
  adaptation.
\newblock In \emph{ICML}, 2019.

\bibitem[Zellinger et~al.(2021)Zellinger, Shepeleva, Dinu, Eghbal-zadeh,
  Nguyen, Nessler, Pereverzyev, and Moser]{Balancing}
Werner Zellinger, Natalia Shepeleva, Marius-Constantin Dinu, Hamid
  Eghbal-zadeh, Duc~Hoan Nguyen, Bernhard Nessler, Sergei~V. Pereverzyev, and
  Bernhard~Alois Moser.
\newblock The balancing principle for parameter choice in distance-regularized
  domain adaptation.
\newblock In \emph{NeurIPS}, 2021.

\bibitem[Zhang et~al.(2022)Zhang, Li, Liu, Zhang, Su, Zhu, shuan Ni, and yeung
  Shum]{DINO}
Hao Zhang, Feng Li, Siyi Liu, Lei Zhang, Hang Su, Jun-Juan Zhu, Lionel~Ming
  shuan Ni, and Heung yeung Shum.
\newblock Dino: Detr with improved denoising anchor boxes for end-to-end object
  detection.
\newblock \emph{ArXiv}, abs/2203.03605, 2022.

\bibitem[Zhang et~al.(2019)Zhang, Liu, Long, and Jordan]{MDD}
Yuchen Zhang, Tianle Liu, Mingsheng Long, and Michael~I. Jordan.
\newblock Bridging theory and algorithm for domain adaptation.
\newblock In \emph{ICML}, 2019.

\bibitem[Zoph \& Le(2016)Zoph and Le]{NAS}
Barret Zoph and Quoc~V. Le.
\newblock Neural architecture search with reinforcement learning.
\newblock In \emph{ICLR}, 2016.

\bibitem[Zou et~al.(2023)Zou, Yang, Zhang, Li, Li, Gao, and Lee]{SEEM}
Xueyan Zou, Jianwei Yang, Hao Zhang, Feng Li, Linjie Li, Jianfeng Gao, and
  Yong~Jae Lee.
\newblock Segment everything everywhere all at once.
\newblock \emph{ArXiv}, abs/2304.06718, 2023.

\end{thebibliography}

\clearpage
\appendix

\section{Implementations of Unsupervised Domain Adaptation Metrics}

\begin{table*}[h]
    \centering
    \resizebox{1\textwidth}{!}{
        \begin{tabular}{c|c|c|c}
            \toprule
            Metric & w. Source Accuracy & Hard to Attack & Input-level \\
            \hline
            $\mathcal{A}$-distance\cite{BenDavidA} & \checkmark & \checkmark & \\
            $\mathcal{H} \Delta \mathcal{H}$-divergence or MCD~\cite{BenDavidH,MCD} & \checkmark & \checkmark & \\
            MDD~\cite{MDD} & \checkmark & \checkmark & \\
            \hline
            Deep Embedded Validation (DEV)~\cite{DEV} & \checkmark & & \\
            DEVN~\cite{ThreeNV} & \checkmark & & \\
            Entropy ~\cite{SSLEntropy, ADVENT} & & \\ 
            Soft Neighborhood Density (SND)~\cite{SND} & & \\ 
            Mutual Information ~\cite{MI} & & \\ 
            BNM~\cite{ThreeNV} & & \\ 
            ClassAMI~\cite{ThreeNV} & & \\
            \hline
            ISM (ours) & \checkmark & \checkmark &\\
            ACM (ours) & \checkmark & \checkmark& \checkmark\\
            \bottomrule
        \end{tabular}}
    \caption{The metrics of UDA studied in the paper. We implement previous metrics according to their papers and modify them to be positively correlated with target accuracy.}
    \label{tab:metric}
\end{table*}

\subsection{Discrepancy-based Metric:} 

Ben-David's theory~\cite{BenDavidA,BenDavidH} shows that the error rate of a classifier on the target domain can be bounded by the error rate on the source domain and the domain divergence:
\begin{align}
    \epsilon_T(h) \leq \epsilon_S(h)+d_{\mathcal{H}}\left(\mathcal{D}_S, \mathcal{D}_T\right)+\lambda
\end{align}
where $\lambda=\lambda_T+\lambda_S$, and $\lambda_T$ and $\lambda_S$ are the errors of $h^*=\operatorname{argmin}_{h \in H}\left(\epsilon_T(h)+\epsilon_S(h)\right)$ with respect
to $\mathcal{D}_T$ and $\mathcal{D}_S$ respectively. Later works~\cite{DANN} exploits this bound to optimize the domain divergence and source error to minimize the target error. Inspired by this formula, we think that the target error can be approximated by domain divergence and source error. In other words, we can utilize domain divergence and source accuracy as the evaluation metric to measure target accuracy.

We transform these discrepancy-based UDA methods ~\cite{BenDavidA,BenDavidH,MCD,MDD} into UDA metrics. These metrics are composited by source accuracy and domain divergence. We can formalize the UDA metrics as:
\begin{align}
    \mathcal{M}(\mathcal{D}_S, \mathcal{D}_T, \boldsymbol{M}) = 
    A_S(\mathcal{D}_S, \boldsymbol{M}) - d_{\mathcal{M}}(\mathcal{D}_S, \mathcal{D}_T, \boldsymbol{M})
\end{align}
where $\boldsymbol{M}$ is the model to be evaluated, $ A_S$ is source accuracy and $d_{\mathcal{M}}$ is the domain divergence. The model is composed of a feature generator and a classifier: $\boldsymbol{M}=\textbf{f}(\textbf{g}(\cdot))$.
Different metrics for UDA have different $d_{\mathcal{M}}$ terms, we describe each $d_{\mathcal{M}}$ terms in the following.

\textbf{1) $\mathcal{A}$-distance}~\cite{BenDavidA}:
\begin{align}
    d_{\mathcal{A}}=&2\sup_{h \in \mathcal{H}}|\mathbb{E}_{\mathcal{D}_s} I\left[h=1\right]+\mathbb{E}_{\mathcal{D}_t} I\left[h=0\right]|\nonumber
\end{align}

A domain discriminator $h$ is trained and the accuracy of the domain discriminator is used as the metric. We use one linear layer to model the domain discriminator the same as~\cite{tllib}.
Notably, when evaluating metrics, we only have the validation set of the source and the target domain, but we need to train the domain discriminator on a training set and evaluate it on the other set. So we use 3-fold validation: we split the validation set into three parts, and each time we train the domain discriminator on two parts and evaluate it on the left part. If not specified, for the following metric that needs training additional networks, we use this 3-fold validation to get the metric score.

\textbf{2) $\mathcal{H} \Delta \mathcal{H}$-divergence or MCD}~\cite{BenDavidH,MCD}:
\begin{align}
    d_{\mathcal{H} \Delta \mathcal{H}}&=\sup _{h, h^{\prime} \in \mathcal{H}}\left|\mathbb{E}_{\mathcal{D}_s} I\left[h^{\prime} \neq h\right]-\mathbb{E}_{\mathcal{D}_t} I\left[h^{\prime} \neq h\right]\right|\nonumber
\end{align}

Two additional classifiers are trained on the top of the feature. Apart from supervised training on source features, they also need to agree on the source domain and disagree on the target domain. These two classifiers are modeled by one linear layer. 

\textbf{3) Maximum Mean Discrepancy (MDD)}~\cite{MDD}:
\begin{align}
    d_{f, \mathcal{F}}^{(\rho)}({\mathcal{D}_s}, {\mathcal{D}_t}) = \sup _{f^{\prime}\in\mathcal{F}}\left(\operatorname{disp}_{\mathcal{D}_s}^{(\rho)}\left(f, f^{\prime}\right)-\operatorname{disp}_{\mathcal{D}_t}^{(\rho)}\left(f, f^{\prime}\right)\right)\nonumber
\end{align}
where $f$ is the classifier of the evaluated model and $\operatorname{disp}^{(\rho)}$ is the margin error. An additional classifier $f'$ is trained to agree $f$ on the source domain and disagree $f$ on the target domain. $f'$ is modeled by a linear layer and trained by the algorithm: Eq. (30) in the original paper~\cite{MDD}.

\subsection{Importance Weighted Validation Metric:} 

\textbf{4) Deep Embedded Validation (DEV)}~\cite{DEV}:

DEV is based on the Importance-Weighted cross-validation (IWCV) of the source domain. It needs to train a two-layer domain discriminator $h$ first, and compute IWCV:
\begin{align}
    \ell(\mathbf{x}_i^{s}) &= w_{h}\left(\mathbf{x}_i^{s}\right) I\left(\hat{y}_i^{s}\neq y_i^{s}\right)\nonumber \\
    w_{h}\left(\mathbf{x}_i^{s}\right) &= \frac{n_s}{n_t} \frac{1-h\left(\boldsymbol{z}_i^{s}\right)}{h\left(\boldsymbol{z}_i^{s}\right)}\nonumber
\end{align}
Then DEV adds IWCV and the variance of the risk estimation as follows:
\begin{align}
    \boldsymbol{DEV}&=\operatorname{mean}(\ell)+\eta \operatorname{mean}(W)-\eta \\
    \eta&=-\frac{\widehat{\operatorname{Cov}}\left(\ell, w_h\right)}{\widehat{\operatorname{Var}}\left[w_h\right]}
\end{align}
We use the negative DEV to be positively related to accuracy.

\textbf{5) DEV with normalization (DEVN)}~\cite{ThreeNV}:

In \cite{ThreeNV}, they propose to normalize the weights by either max normalization or standardization to avoid large $\eta$. We implement DEVN with standardization:
\begin{align}
W_{s t}=\frac{W-\bar{W}}{\sigma_W}+1
\end{align}

Then $W_{s t}$ is used in $\boldsymbol{DEV}$.

\subsection{Entropy-based Metric:} 

\textbf{6) Entropy}~\cite{C-Ent,ADVENT}:
\begin{align}
    \boldsymbol{Ent} = -\mathbb{E}_{\mathcal{D}_t}[H(\boldsymbol{p})]=\mathbb{E}_{\mathcal{D}_t}[\sum_k \boldsymbol{p}_k\log \boldsymbol{p}_k]\nonumber
\end{align}

We compute the negative entropy of the predicted probability $\boldsymbol{p}$ of $\boldsymbol{M}$ on target samples.

\textbf{7) Soft Neighborhood Density
(SND)~\cite{SND}}:

In their work, they define the soft neighborhoods as the similarity distribution between target samples and estimate the density by computing the entropy of the distribution. The similarity is defined as $S_{i j}=\left\langle\boldsymbol{p}_i^t, \boldsymbol{p}_j^t\right\rangle$. The similarity distribution is computed as follows:

\begin{align}
P_{i j}&=\frac{\exp \left(S_{i j} / \tau\right)}{\sum_{j^{\prime}} \exp \left(S_{i j^{\prime}} / \tau\right)}
\end{align}
Then SND is defined as:
\begin{align}
\boldsymbol{SND}&=-\frac{1}{N_t} \sum_{i=1}^{N_t} \sum_{j=1}^{N_t} P_{i j} \log P_{i j}
\end{align}

\textbf{8) Mutual Information~\cite{MI}}:

The Mutual Information of the model prediction:
\begin{align}
    \boldsymbol{MI} = H(\mathbb{E}_{\mathcal{D}_t}[\boldsymbol{p}])-\mathbb{E}_{\mathcal{D}_t}[H(\boldsymbol{p})]\nonumber
\end{align}

\subsection{Other Metric:} 

\textbf{9) Batch nuclear-norm maximization (BNM)~\cite{BNM,ThreeNV}}:

BNM is a UDA algorithm that aims to generate diverse and confident predictions. It approaches this via singular value decomposition:
\begin{equation}
\boldsymbol{BNM}=\|P\|_*
\end{equation}
where $P$ is the $\tilde{N}_t \times K$ prediction matrix ($\tilde{N}_t$ is the target validation set size, and $K$ is the number of classes), and $\|P\|_*$ is the nuclear norm (the sum of the singular values) of $P$.

\textbf{10) ClassAMI~\cite{ThreeNV}}:

They propose computing the Adjusted Mutual Information (AMI) between target cluster labels and the predicted labels:
\begin{align}
\boldsymbol{ClassAMI}&=\operatorname{AMI}(P, \operatorname{kmeans}(F)_{\cdot} \text{labels}) \\
P_i &= \underset{k}{\operatorname{argmax}}
[\boldsymbol{p}_i]
\end{align}
where $P$ is the predicted labels for the target data, $\boldsymbol{p}_i$ is the i-th prediction vector, and $F$ is the set of target features.

\subsection{Our Metric:} 

\textbf{11) Inception Score Metric for UDA (ISM)}:

Its formula is presented in the main paper. The MLP classifier $\boldsymbol{h}$ has two layers with a hidden size equal to the feature size (bottleneck dimension). The classifier is trained by the LBFGS optimizer for 200 steps on the source validation set.

\textbf{12) Augmentation Consistency Metric (ACM)}:

Its formula is presented in the main paper. For the data-augmented sample, we use a series of random data augmentation to get it, including Random Resize and Crop, Horizontal Flip, Random Color Jitter, and Random Gaussian Blur.

\begin{table}[t]
    \centering
    \begin{tabular}{ccccc}
        \toprule
        Office31 & A & W & D &     \\
        \hline
        Training & 1,971 & 556 & 498 & \\
        Validation & 846 & 239 & 498 & \\
        \hline
        OfficeHome & Ar & Cl & Pr & Rw     \\
        \hline
        Training & 1,698 & 3,055 & 3,107 & 3,049 \\
        Validation & 729 & 1,310 & 1,332 & 1,308 \\
        \hline
        DomainNet & c & p & r & s     \\
        \hline
        Training & 33,525 & 50,416 & 120,906 & 48,212 \\
        Validation & 14,604 & 21,850 & 52,041 & 20,916 \\
        \hline
        VisDA & Syn & Real & &    \\
        \hline
        Training & 106,677 & 55,388 & & \\
        Validation & 45,720 & 72,372 & & \\
        \bottomrule
    \end{tabular}
    \caption{The statistic of the training set and the validation set of the datasets used in the paper. Following the 70\%/30\% scheme, we split Office31, OfficeHome, and the ``Synthetic'' domain of VisDA into no overlapping training and validation sets.}
    \label{tab:datasets}
\vspace{-2mm}
\end{table}

\section{Details of UDA training}
\subsection{Datasets Splitting}
Generally speaking, the validation set contains different samples with the training set to show generalization. However, Office31 and OfficeHome do not split the training and validation set, so previous works report the accuracy of the target training set. To solve this historical issue, we follow a 70\%/30\% split scheme to split the training and the validation set for Office31 and OfficeHome (except for ``D'' domain of Office31, due to limited samples) and report the target accuracy on the validation set. We also split the training and validation set for the source domain of VisDA2017, as metrics will utilize the source validation set. The training sets and validation sets are listed in Tab~\ref{tab:datasets}

\subsection{Training Implement Details}
Following Transfer-Learning-Library~\cite{tllib}, we train all five methods (Source only, DANN, CDAN, MDD, MCC) through SGD with $0.9$ momentum, and the learning rate of ResNet backbone is scaled by $0.1$. We schedule the learning rate with the commonly used strategy: the learning rate is adjusted by $\eta_p=\frac{\eta_0}{(1+\alpha q)^{\beta}}$,  where $q$ is the training progress linearly changing from $0$ to $1$, $\eta_0=0.01$, $\alpha=10$, $\beta=0.75$. The batch size is set to $32$ for all training. We train each model with one V100 GPU. For the architecture of the model, $\boldsymbol{M}=\textbf{f}(\textbf{g}(\cdot))$, the feature generator $\textbf{g}$ contain a ResNet~\cite{ResNet} backbone and a bottleneck layer, and the classifier $\textbf{f}$ is a linear layer. We use ResNet50 as the backbone for Office31 and OfficeHome, and ResNet101 for VisDA and DomainNet. We train every model for 3000 steps on Office31, OfficeHome, and VisDA, and 6000 steps on DomainNet in total. 

\begin{figure}[htbp]
\begin{minipage}[t]{1\linewidth}
    \centering
    \resizebox{1\textwidth}{!}{  
        \begin{tabular}{c|a|a|a|a|a|a}
            \toprule
            Datasets & \multicolumn{4}{|c}{OfficeHome} & \multicolumn{4}{|c}{VisDA2017} & \multicolumn{4}{|c}{DomainNet} \\
            \hline
            Train Method & \multicolumn{2}{|c}{MI} & \multicolumn{2}{|c}{AC} & \multicolumn{2}{|c}{MI} & \multicolumn{2}{|c}{AC} & \multicolumn{2}{|c}{MI} & \multicolumn{2}{|c}{AC} \\
            \hline
            Metric & corr & dev & corr & dev & corr & dev & corr & dev & corr & dev & corr & dev \\
            \hline
            MI & 0.67 & 7.97 & 0.55 & 5.76 & - 0.21 & 15.3 & 0.87 & 1.99 & -0.75 & 18.9 & 0.95 & 1.62 \\
            ISM & 0.89 & 1.73 & \textcolor{red}{\textbf{0.97}} & 1.43 & 0.87 & 1.77 & 0.89 & 1.57 & 0.88 & \textcolor{red}{\textbf{0.0}} & \textcolor{red}{\textbf{0.97}} & \textcolor{red}{\textbf{0.62}} \\
            ACM & \textcolor{red}{\textbf{0.92}} & \textcolor{red}{\textbf{1.15}} & 0.62 & \textcolor{red}{\textbf{1.37}} & \textcolor{red}{\textbf{0.99}} & \textcolor{red}{\textbf{0.0}} & \textcolor{red}{\textbf{0.89}} & \textcolor{red}{\textbf{0.59}} & \textcolor{red}{\textbf{0.91}} & \textcolor{red}{\textbf{0.0}} & 0.96 & 1.55 \\
            \bottomrule
    \end{tabular}}
    \captionof{table}{A test of whether metrics are attackable on different datasets. These three metrics are transformed into training loss, and then trained models are evaluated by themselves.}
    \label{tab:robustness}
\vspace{2mm}
\end{minipage}

\begin{minipage}[t]{1\linewidth}
    \centering
    \resizebox{1\textwidth}{!}{
    \begin{tabular}{c|a|a|a|a|a|a}
        \toprule
        Training Method & \multicolumn{2}{|c}{Source only} & \multicolumn{2}{|c}{DANN} & \multicolumn{2}{|c}{CDAN} & \multicolumn{2}{|c}{MDD} & \multicolumn{2}{|c}{MCC} & \multicolumn{2}{|c}{ALL} \\
        \hline
        Metric & corr & dev & corr & dev & corr & dev & corr & dev & corr & dev & corr & dev \\
        \hline
        MDD & -0.65 & 7.39 & 0.58 & 6.29 & -0.11 & 4.73 & 0.66 & 0.85 & 0.34 & 4.38 & 0.36 & 5.93 \\
        DEVN & -0.6 & 7.39 & -0.16 & 6.67 & -0.15 & 8.99 & 0.55 & 4.97 & 0.47 & 4.99 & 0.37 & 32.51 \\
        BNM & 0.11 & 6.35 & 0.60 & 3.65 & -0.03 & 3.72 & \textcolor{blue}{0.85} & \textcolor{red}{\textbf{0.00}} & 0.02 & 3.41 & 0.48 & 3.41 \\
        ClassAMI & -0.40 & 7.39 & 0.74 & 6.29 & -0.12 & 9.81 & \textcolor{red}{\textbf{0.94}} & 1.29 & 0.15 & 7.19 & 0.61 & 7.76 \\
        \hline
        ISM & \textcolor{red}{\textbf{0.84}} & \textcolor{red}{\textbf{0.31}} & \textcolor{blue}{0.75} & \textcolor{blue}{3.92} & \textcolor{blue}{0.42} & \textcolor{blue}{1.23} & 0.75 & \textcolor{blue}{0.40} & \textcolor{blue}{0.88} & \textcolor{red}{\textbf{0.66}} & \textcolor{blue}{0.59} & \textcolor{red}{\textbf{1.66}} \\
        ACM & \textcolor{blue}{0.80} & \textcolor{blue}{2.38} & \textcolor{red}{\textbf{0.79}} & \textcolor{red}{\textbf{1.18}} & \textcolor{red}{\textbf{0.61}} & \textcolor{red}{\textbf{0.98}} & \textcolor{blue}{0.85} & \textcolor{red}{\textbf{0.0}} & \textcolor{red}{\textbf{0.93}} & \textcolor{blue}{1.66} & \textcolor{red}{\textbf{0.76}} & \textcolor{red}{\textbf{1.66}} \\
        \bottomrule
    \end{tabular}}
    \caption{Consistency between metrics of UDA and target accuracy on VisDA2017.}
    \label{tab:visda2}
\vspace{2mm}
\end{minipage}

\begin{minipage}[t]{1\linewidth}
    \centering
    \resizebox{1\textwidth}{!}{
    \begin{tabular}{c|a|a|a|a|a|a}
        \toprule
        Training Method & \multicolumn{2}{|c}{Source only} & \multicolumn{2}{|c}{DANN} & \multicolumn{2}{|c}{CDAN} & \multicolumn{2}{|c}{MDD} & \multicolumn{2}{|c}{MCC} & \multicolumn{2}{|c}{ALL} \\
        \hline
        Metric & corr & dev & corr & dev & corr & dev & corr & dev & corr & dev & corr & dev \\
        \hline
        MDD & 0.53 & 4.44 & 0.72 & 1.67 & 0.83 & 1.88 & 0.83 & \textcolor{blue}{1.13} & 0.05 & 5.99 & 0.3 & 6.2 \\
        DEVN & 0.18 & 3.48 & -0.06 & 6.38 & 0.08 & 7.10 & 0.89 & 12.03 & 0.27 & 8.45 & 0.52 & 10.14 \\
        BNM & -0.59 & 6.63 & 0.48 & 4.53 & 0.88 & 1.30 & 0.93 & 1.98 & 0.37 & 17.19 & 0.54 & 17.28 \\
        ClassAMI & \textcolor{red}{\textbf{0.79}} & 3.57 & \textcolor{red}{\textbf{0.77}} & 2.93 & 0.86 & 1.25 & 0.93 & \textcolor{blue}{1.08} & \textcolor{blue}{0.71} & \textcolor{red}{\textbf{1.16}} & 0.81 & \textcolor{red}{\textbf{1.24}} \\
        \hline
        ISM & 0.72 & \textcolor{blue}{1.47} & 0.6 & \textcolor{blue}{1.68} & \textcolor{red}{\textbf{0.91}} & \textcolor{blue}{1.15} & \textcolor{red}{\textbf{0.97}} & 1.45 & 0.70 & 1.96 & \textcolor{blue}{0.88} & 1.96 \\
        ACM & \textcolor{blue}{0.75} & \textcolor{red}{\textbf{1.37}} & \textcolor{red}{\textbf{0.77}} & \textcolor{red}{\textbf{1.16}} & \textcolor{blue}{0.90} & \textcolor{red}{\textbf{1.13}} & \textcolor{blue}{0.95} & \textcolor{red}{\textbf{0.93}} & \textcolor{red}{\textbf{0.94}} & \textcolor{blue}{1.36} & \textcolor{red}{\textbf{0.93}} & \textcolor{blue}{1.73} \\
        \bottomrule
    \end{tabular}}
    \captionof{table}{The ``corr'' and ``dev'' results are averaged over the 12 transfer tasks of OfficeHome. }
    \label{tab:officehome2}
\vspace{2mm}
\end{minipage}

\begin{minipage}[t]{1\linewidth}
    \centering
    \resizebox{1\textwidth}{!}{
    \begin{tabular}{c|a|a|a|a|a|a}
        \toprule
        Training Method & \multicolumn{2}{|c}{Source only} & \multicolumn{2}{|c}{DANN} & \multicolumn{2}{|c}{CDAN} & \multicolumn{2}{|c}{MDD} & \multicolumn{2}{|c}{MCC} & \multicolumn{2}{|c}{ALL} \\
        \hline
        Metric & corr & dev & corr & dev & corr & dev & corr & dev & corr & dev & corr & dev \\
        \hline
        MDD & 0.26 & 2.12 & 0.54 & 3.46 & 0.72 & 1.11 & 0.7 & 12.75 & -0.55 & 10.39 & 0.54 & 4.99 \\
        DEVN & 0.81 & 1.47 & 0.67 & 5.40 & 0.80 & 7.99 & 0.87 & 0.38 & 0.50 & 9.95 & 0.0 & 2.45 \\
        BNM & 0.34 & 3.83 & 0.58 & 6.65 & 0.65 & 4.28 & 0.34 & 14.28 & 0.74 & 1.37 & 0.59 & 2.85 \\
        ClassAMI & 0.57 & \textcolor{red}{\textbf{1.27}} & 0.52 & 3.46 & 0.60 & 5.62 & 0.61 & \textcolor{red}{\textbf{0.04}} & 0.81 & 1.37 & 0.65 & 3.55 \\
        \hline
        ISM & \textcolor{blue}{0.85} & \textcolor{blue}{1.33} & \textcolor{red}{\textbf{0.87}} & \textcolor{blue}{0.7} & \textcolor{blue}{0.96} & \textcolor{blue}{0.41} & \textcolor{red}{\textbf{0.98}} & \textcolor{blue}{0.28} & \textcolor{red}{\textbf{0.92}} & \textcolor{blue}{0.6} & \textcolor{red}{\textbf{0.91}} & \textcolor{blue}{1.13} \\
        ACM & \textcolor{red}{\textbf{0.94}} & 1.41 & \textcolor{blue}{0.8} & \textcolor{red}{\textbf{0.27}} & \textcolor{red}{\textbf{0.98}} & \textcolor{red}{\textbf{0.29}} & \textcolor{blue}{0.93} & \textcolor{red}{\textbf{0.04}} & \textcolor{blue}{0.84} & \textcolor{red}{\textbf{0.15}} & \textcolor{blue}{0.87} & \textcolor{red}{\textbf{0.29}} \\
        \bottomrule
    \end{tabular}}
    \captionof{table}{The ``corr'' and ``dev'' results are averaged over 12 transfer tasks of DomainNet. }
    \label{tab:domainnet2}
\vspace{2mm}
\end{minipage}

\begin{minipage}[t]{1\linewidth}
    \centering
    \resizebox{0.6\textwidth}{!}{  
        \begin{tabular}{c|a|a|a}
            \toprule
            Training Method & \multicolumn{2}{|c}{DANN} & \multicolumn{2}{|c}{CDAN} & \multicolumn{2}{|c}{ALL} \\
            \hline
            Metric & corr & dev & corr & dev & corr & dev \\
            \hline
            $\mathcal{A}$-distance & 0.37 & 1.69 & 0.34 & 1.3 & 0.35 & 3.11 \\
            MCD & 0.46 & \textcolor{blue}{1.48} & 0.32 & 1.71 & 0.48 & 2.20 \\
            MDD & 0.53 & 2.46 & 0.38 & 1.34 & 0.63 & 2.22 \\
            DEV & NaN & - & NaN & -  & NaN & - \\
            DEVN & NaN & - & NaN & -  & NaN & - \\
            Entropy & 0.4 & 2.47 & 0.57 & 2.59 & 0.55 & 2.01 \\
            SND & 0.43 & 6.70 & 0.44 & 3.09 & 0.57 & 6.14 \\
            MI & 0.38 & 2.26 & 0.58 & 1.51 & 0.53 & 2.74 \\
            BNM & 0.29 & 2.99 & 0.54 & 1.75 & 0.32 & 3.59 \\
            ClassAMI & 0.56 & 1.83 & \textcolor{blue}{0.61} & 2.06 & 0.67 & 3.46 \\
            \hline
            ISM & \textcolor{red}{\textbf{0.73}} & 1.52 & \textcolor{red}{\textbf{0.63}} & \textcolor{red}{\textbf{1.04}} & \textcolor{blue}{0.71} &  \textcolor{red}{\textbf{1.41}} \\
            ACM & \textcolor{blue}{0.71} & \textcolor{red}{\textbf{1.46}} & 0.59 & \textcolor{blue}{1.21} & \textcolor{red}{\textbf{0.75}} & \textcolor{blue}{1.84} \\
            \bottomrule
    \end{tabular}}
    \captionof{table}{Consistency between metrics of UDA and target accuracy on Office31. The results are averaged across 6 transfer tasks of Office31.}
    \label{tab:office31}
\vspace{0mm}
\end{minipage}
\end{figure}

\section{Experimental Results}

\subsection{More Consistency Results}

In Section 4.2 of the main paper, we show the results of comparing our ISM and ACM to previous metrics on three datasets. In the Appendix, we add three more previous metrics~\cite{ThreeNV}: DEVN, BNM, and ClassAMI. We show the consistency of each metric with target accuracy on four UDA datasets with five UDA training methods. Tab.~\ref{tab:visda2}, Tab.~\ref{tab:domainnet2}, Tab.~\ref{tab:officehome2} and Tab.~\ref{tab:office31} show the results on VisDA2017, DomainNet, OfficeHome and Office31 respectively.

We find the performance of the metric can vary largely for different training methods. DEVN, BNM, and ClassAMI demonstrate excellent performance in certain cases, yet they may also exhibit significant errors under other conditions. Our ISM and ACM show decent performance for all training methods on all datasets. We find the results of all metrics decrease on Office31, which may be due to small validation sets. DEV and DEVN will collapse on Office31 because source accuracy can be $1$.

\begin{figure}[htbp]
\begin{minipage}[t]{1\linewidth}
	\centering
	\resizebox{1.0\textwidth}{!}{  
		\begin{tabular}{cccccccccccccc}
			\toprule
			Method & c $\to$ p & c $\to$ r & c $\to$ s & p $\to$ c & p $\to$ r & p $\to$ s & r $\to$ c & r $\to$ p & r $\to$ s & s $\to$ c & s $\to$ p & s $\to$ r & Avg    \\
			\hline
			DANN (default) & 35.9 & 54.3 & 43.8 & 38.0 & 54.9 & 35.5 & 49.8 & 50.1 & 38.3 & 54.4 & 43.8 & 53.2 & 46.0 \\
			DANN (searched) & 38.0 & 54.5 & 44.7 & 40.7 & 56.1 & 37.9 & 50.7 & 50.7 & 38.3 & 55.0 & 44.7 & 53.7 & 47.1\\
			Gains ($+\Delta$) & \textcolor{red}{+2.1} & \textcolor{red}{+0.2} & \textcolor{red}{+0.9} & \textcolor{red}{+2.7} & \textcolor{red}{+1.2} & \textcolor{red}{+2.4} & \textcolor{red}{+0.9} & \textcolor{red}{+0.6} & \textcolor{red}{+0.0} & \textcolor{red}{+0.6} & \textcolor{red}{+0.9} & \textcolor{red}{+0.5} & \textcolor{red}{\textbf{+1.1}}\\
			\hline
			CDAN (default) & 40.0 & 55.8 & 44.6 & 44.2 & 57.3 & 39.8 & 55.2 & 53.3 & 41.5 & 56.9 & 46.3 & 55.5 & 49.2 \\
			CDAN (searched) & 40.6 & 56.5 & 45.1 & 45.5 & 58.4 & 40.3 & 55.4 & 53.1 & 42.3  & 57.1 & 46.6 & 56.4 & 49.8 \\
			Gains ($+\Delta$) & \textcolor{red}{+0.6} & \textcolor{red}{+0.7} & \textcolor{red}{+0.5} & \textcolor{red}{+1.3} & \textcolor{red}{+1.1} & \textcolor{red}{+0.5} & \textcolor{red}{+0.2} & \textcolor{green}{-0.2} & \textcolor{red}{+0.8} & \textcolor{red}{+0.2} & \textcolor{red}{+0.3} & \textcolor{red}{+0.9} & \textcolor{red}{\textbf{+0.6}} \\
			\hline
			MDD (default) & 42.3 & 58.4 & 46.6 & 48.5 & 60.1 & 43.6 & 56.8 & 56.3 & 46.3 & 57.2 & 44.8 & 57.2 & 51.5 \\
			MDD (searched) & 42.5 & 58.6 & 47.0 & 48.5 & 60.1 & 43.6 & 57.3 & 55.9 & 46.6 & 57.5 & 45.0 & 57.2 & 51.7 \\
			Gains ($+\Delta$) & \textcolor{red}{+0.2} & \textcolor{red}{+0.2} & \textcolor{red}{+0.4} & \textcolor{red}{+0.0} & \textcolor{red}{+0.0} & \textcolor{red}{+0.0} & \textcolor{red}{+0.5} & \textcolor{green}{-0.4} & \textcolor{red}{+0.3} & \textcolor{red}{+0.3} & \textcolor{red}{+0.2} & \textcolor{red}{+0.0} & \textcolor{red}{\textbf{+0.2}} \\
			\hline
			MCC (default) & 35.1 & 49.2 & 40.6 & 41.0 & 56.0 & 36.2 & 48.3 & 49.0 & 36.3 & 51.9 & 38.9 & 49.9 & 44.4 \\
			MCC (searched) & 41.2 & 53.6 & 44.5 & 51.1 & 59.9 & 40.7 & 58.5 & 54.8 & 38.2 & 61.7 & 47.6 & 55.0 & 50.6 \\
			Gains ($+\Delta$) & \textcolor{red}{+6.1} & \textcolor{red}{+4.4} & \textcolor{red}{+3.9} & \textcolor{red}{+10.1} & \textcolor{red}{+3.9} & \textcolor{red}{+4.5} & \textcolor{red}{+10.2} & \textcolor{red}{+5.8} & \textcolor{red}{+1.9} & \textcolor{red}{+9.8} & \textcolor{red}{+8.7} & \textcolor{red}{+5.1} & \textcolor{red}{\textbf{+6.2}} \\
			\bottomrule
	\end{tabular}}
	\captionof{table}{The hyper-parameters found by our metric v.s. the default hyper-parameters in original papers on DomainNet.}
	\label{tab:domainnet-search}
\vspace{1mm}
\end{minipage}

\begin{minipage}[t]{1\linewidth}
    \centering
    \resizebox{1.0\textwidth}{!}{  
        \begin{tabular}{clc}
            \toprule
            Method & Ar $\to$ Cl Ar $\to$ Pr Ar $\to$ Rw Cl $\to$ Ar Cl $\to$ Pr Cl $\to$ Rw\,Pr $\to$ Ar Pr $\to$ Cl Pr $\to$ Rw Rw $\to$ Ar Rw $\to$ Cl\,Rw $\to$ Pr  & Avg    \\
            \hline
            DANN (default) &\quad 49.0\qquad 61.3\qquad 72.9\qquad53.5\qquad 66.6\qquad 68.6\qquad 55.0\qquad 50.4\qquad 75.2       \,\,\,\qquad 67.1       \,\qquad 56.3      \,\qquad 79.3& 62.9  \\
            DANN (searched) &\quad 51.2\qquad 62.3\qquad 74.2\qquad56.7\qquad 66.0\qquad 70.8\qquad 58.7\qquad 52.7\qquad 76.0       \,\,\,\qquad 67.5      \,\qquad 57.8      \,\qquad 80.8& 64.5  \\
            Gains ($+\Delta$) &\quad \textcolor{red}{+2.2}\qquad \textcolor{red}{+1.0}\qquad \textcolor{red}{+1.3}  \,\,\,\,\quad \textcolor{red}{+3.2}\qquad \textcolor{green}{-0.6}\qquad \textcolor{red}{+2.2}\qquad \textcolor{red}{+3.7}  \,\,\,\,\quad \textcolor{red}{+2.3}  \,\,\,\,\quad \textcolor{red}{+0.8} \,\,\qquad \textcolor{red}{+0.4} \,\qquad \textcolor{red}{+1.5} \,\qquad \textcolor{red}{+1.5}& \textcolor{red}{\textbf{+1.6}}\\
            \hline
            CDAN (default) &\quad 50.4\qquad 69.4\qquad 73.5\qquad56.7\qquad 69.4\qquad 69.1\qquad 57.3\qquad 50.5\qquad 75.5       \,\,\,\qquad 70.6       \,\qquad 55.8      \,\qquad 80.6& 64.9  \\
            CDAN (searched) &\quad 51.1\qquad 69.2\qquad 74.3\qquad58.4\qquad 70.3\qquad 69.7\qquad 61.6\qquad 50.6\qquad 77.5       \,\,\,\qquad 71.4       \,\qquad 56.7      \,\qquad 81.1& 66.0 \\
            Gains ($+\Delta$) &\quad \textcolor{red}{+0.7}\qquad \textcolor{green}{-0.3}\qquad \textcolor{red}{+0.8}  \,\,\,\,\quad \textcolor{red}{+1.7}  \,\,\,\,\quad \textcolor{red}{+0.7}\qquad \textcolor{red}{+0.6}\,\,\,\,\,\quad \textcolor{red}{+4.3}  \,\,\,\,\quad \textcolor{red}{+0.1}  \,\,\,\,\quad \textcolor{red}{+2.0} \,\,\qquad \textcolor{red}{+0.8} \,\qquad \textcolor{red}{+0.9} \,\qquad \textcolor{red}{+0.5}& \textcolor{red}{\textbf{+1.1}} \\
            \hline
            MDD (default) &\quad 51.1\qquad 70.6\qquad 72.1\qquad57.3\qquad 70.6\qquad 76.6\qquad 59.5\qquad 53.9\qquad 74.9       \,\,\,\qquad 70.5       \,\qquad 58.6      \,\qquad 81.7& 66.4  \\
            MDD (searched) &\quad 52.9\qquad 72.2\qquad 75.2\qquad58.8\qquad 71.9\qquad 76.6\qquad 58.7\qquad 52.4\qquad 76.8      \,\,\,\qquad 69.8       \,\qquad 59.2      \,\qquad 81.8& 67.2 \\
            Gains ($+\Delta$) & \quad \textcolor{red}{+1.8}\qquad \textcolor{red}{+1.6}\qquad \textcolor{red}{+3.1}  \,\,\,\,\quad \textcolor{red}{+1.5}\qquad \textcolor{red}{+1.3}\qquad \textcolor{red}{+0.0}\qquad \textcolor{green}{-0.8} \,\,\,\,\,\,\quad \textcolor{green}{-1.5} \,\,\,\,\,\quad \textcolor{red}{+1.9} \,\,\,\qquad \textcolor{green}{-0.7} \,\qquad \textcolor{red}{+0.6} \,\qquad \textcolor{red}{+0.1} & \textcolor{red}{\textbf{+0.8}}\\
            \hline
            MCC (default) &\quad 55.5\qquad 77.7\qquad 80.2\qquad62.8\qquad 75.2\qquad 75.8\qquad 61.7\qquad 50.6\qquad 78.3      \,\,\,\qquad 69.7       \,\qquad 56.3      \,\qquad 83.4& 68.9  \\
            MCC (searched) &\quad 56.1\qquad 78.5\qquad 79.0\qquad63.6\qquad 75.2\qquad 76.6\qquad 64.1\qquad 52.3\qquad 78.3       \,\,\,\qquad 71.6       \,\qquad 56.1      \,\qquad 83.5& 69.5 \\
            Gains ($+\Delta$) &\quad \textcolor{red}{+0.6}\,\qquad \textcolor{red}{+0.8}\qquad \textcolor{green}{-1.2}  \,\,\,\,\quad \textcolor{red}{+0.8}\qquad \textcolor{red}{+0.0}\,\,\,\,\,\quad \textcolor{red}{+0.8}\,\,\,\,\,\quad \textcolor{red}{+2.4}  \,\qquad \textcolor{red}{+1.7}  \,\,\,\,\quad \textcolor{red}{+0.0} \,\,\qquad \textcolor{red}{+1.9} \,\,\qquad \textcolor{green}{-0.2} \,\qquad \textcolor{red}{+0.1}& \textcolor{red}{\textbf{+0.6}} \\
            \bottomrule
    \end{tabular}}
    \captionof{table}{The hyper-parameters found by our metric v.s. the default hyper-parameters in original papers on OfficeHome. The target accuracy of 12 transfer tasks is reported.}
    \label{tab:officehome-search}
\vspace{1mm}
\end{minipage}

\begin{minipage}[t]{1\linewidth}
	\centering
        \resizebox{0.8\textwidth}{!}{  
	\begin{tabular}{cccccccc}  
		\toprule
		Method & A $\to$ W & A $\to$ D & W $\to$ A & W $\to$ D & D $\to$ A & D $\to$ W & Avg    \\
		\hline
		DANN (default) & 90.4 & 81.7 & 69.6 & 97.8 & 72.3 & 93.7 & 84.3 \\
		DANN (searched) & 90.6 & 83.9 & 69.6 & 98.6 & 72.3 & 95.0 & 85.1 \\
		Gains ($+\Delta$) & \textcolor{red}{+0.2} & \textcolor{red}{+2.2} & \textcolor{red}{+0.0} & \textcolor{red}{+0.8} & \textcolor{red}{+0.0} & \textcolor{red}{+1.3} & \textcolor{red}{\textbf{+0.8}} \\
		\hline
		CDAN (default) & 91.2 & 93.0 & 68.2 & 100.0 & 72.1 & 97.1 & 86.9 \\
		CDAN (searched) & 91.6 & 91.5 & 69.6 & 100.0 & 74.1 & 97.5 & 87.4 \\
		Gains ($+\Delta$) & \textcolor{red}{+0.4} & \textcolor{green}{-1.5} & \textcolor{red}{+1.4} & \textcolor{red}{+0.0} & \textcolor{red}{+2.0} & \textcolor{red}{+0.4} & \textcolor{red}{\textbf{+0.5}} \\
		\bottomrule
	\end{tabular}}
	\captionof{table}{The hyper-parameters found by our metric v.s. the default hyper-parameters in original papers on Office31.}
	\label{tab:office31-search}
\end{minipage}
\end{figure}

\subsection{Robustness Results}

For the ``Robustness'' property of metrics, we transform MI and ACM into two training methods. We employ these metrics to select the trade-off $\lambda$ from \{0.1, 0.3, 1.0, 3.0, 10.0\} for these methods. We show the implementation of these methods here. The loss of the ``Mutual Information'' method is as follows:
\begin{align}
 L_{MI} = \mathbb{E}_{(\boldsymbol{x}^s,y^s)}[-\log \boldsymbol{p}_{y^s}]+\lambda (\sum_k \hat{\boldsymbol{p}}_k \log \hat{\boldsymbol{p}}_k-\mathbb{E}_{\boldsymbol{x}^t}[\sum_k \boldsymbol{p}_k \log \boldsymbol{p}_k]),
\end{align}
where $\hat{\boldsymbol{p}}_k$ is the average prediction for class $k$ within a batch. 

The loss of the ``Augment Consist'' method is as follows:
\begin{align}
 L_{AC} = \mathbb{E}_{(\boldsymbol{x}^s,y^s)}[-\log \boldsymbol{p}_{y^s}]- \lambda \mathbb{E}_{\boldsymbol{x}^t} \sum_k I[k=\underset{k}{\operatorname{argmax}}(\boldsymbol{p}^t)] \log \boldsymbol{p}^{t \prime} \nonumber
\end{align}
where $\boldsymbol{p}^{t \prime}$ is the prediction of the model on the sample $\boldsymbol{x}^{t \prime}$, which is the augmented version of $\boldsymbol{x}^t$. We use the same random data augmentation as the ``ACM'' metric.

We show the study of the Robustness property on OfficeHome, VisDA2017, and DomainNet in Tab~\ref{tab:robustness} for ``MI'', ``ISM'' and ``ACM'' metrics. When the models are trained with the ``MI'' method, the ``MI'' metric is inconsistent with the target accuracy. Meanwhile, the ``ISM'' metric is robust to this attack. ``ACM'' is also robust against the attack against it.

\begin{figure}[t]
    \centering
    \subfloat[DANN on OfficeHome]{
        \includegraphics[width=0.24\textwidth]{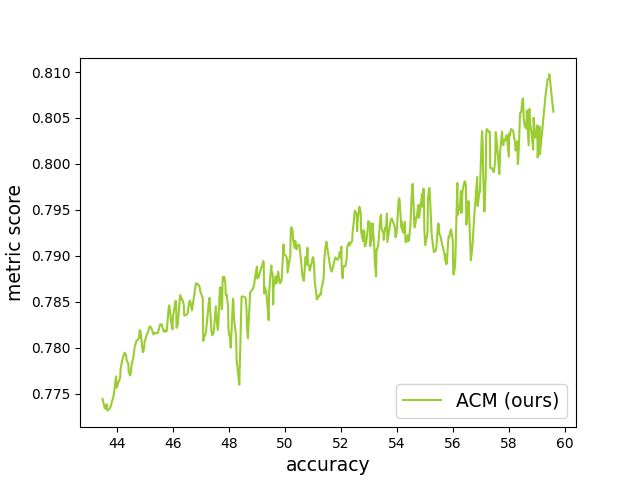}
    }
    \subfloat[CDAN on OfficeHome]{
        \includegraphics[width=0.24\textwidth]{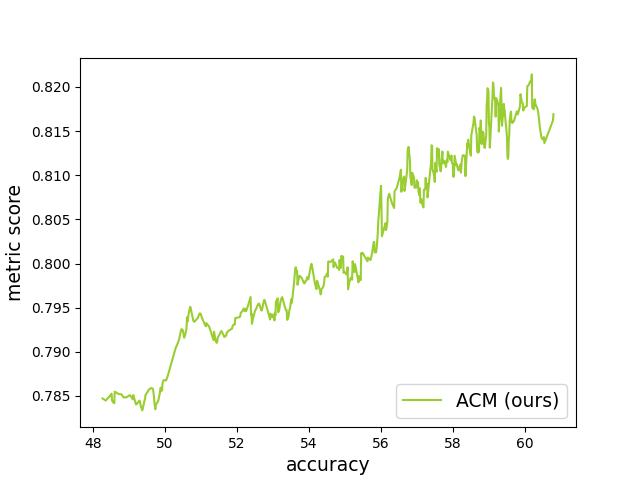}
    }
    \subfloat[DANN on Office31]{
        \includegraphics[width=0.24\textwidth]{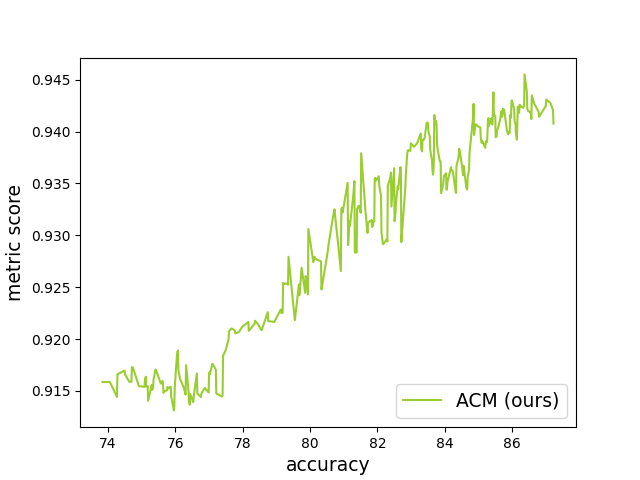}
    }
    \subfloat[CDAN on Office31]{
        \includegraphics[width=0.24\textwidth]{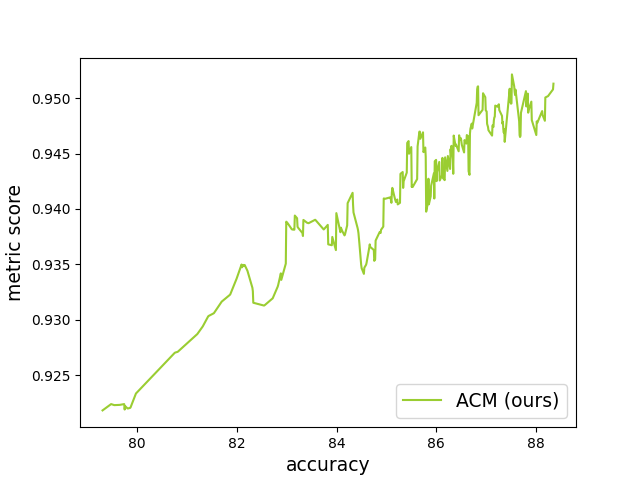}
    }
    \caption{The visualization of the relation between the ACM score and target accuracy. Each sub-figure contains models trained by DANN or CDAN methods on OfficeHome or Office31 datasets.}
    \label{fig:visual_1}
\vspace{-2mm}
\end{figure}

\begin{figure}[t]
    \centering
    \subfloat[DANN on DomainNet]{
        \includegraphics[width=0.24\textwidth]{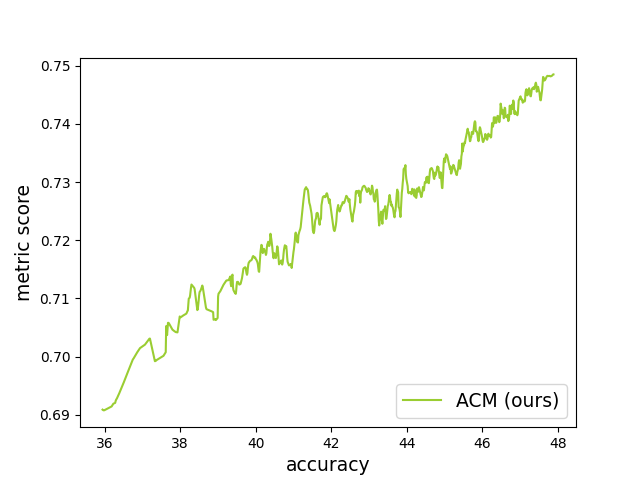}
    }
    \subfloat[CDAN on DomainNet]{
        \includegraphics[width=0.24\textwidth]{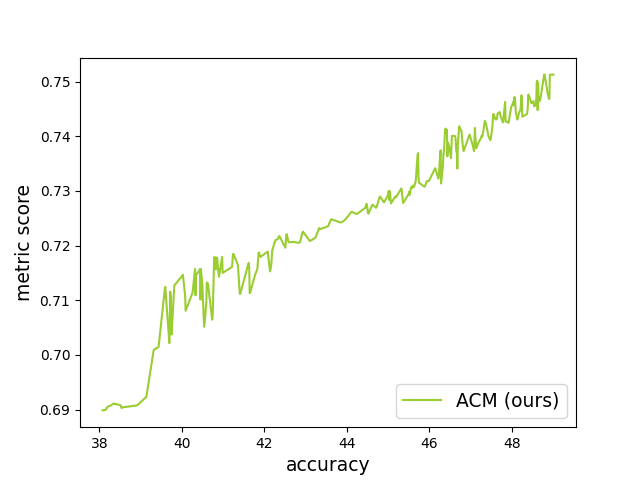}
    }
    \subfloat[DANN on VisDA]{
        \includegraphics[width=0.24\textwidth]{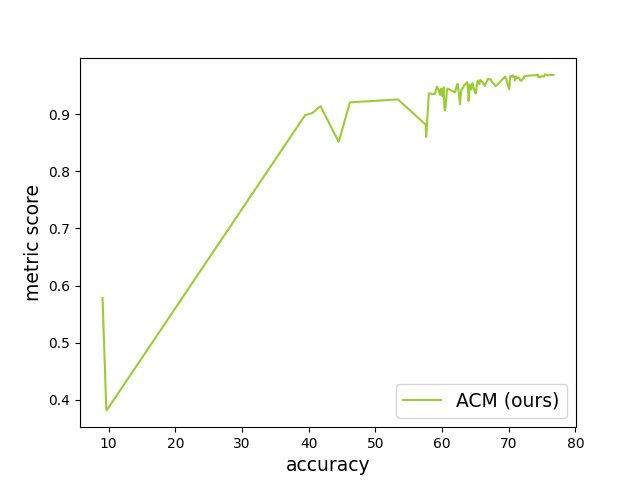}
    }
    \subfloat[CDAN on VisDA]{
        \includegraphics[width=0.24\textwidth]{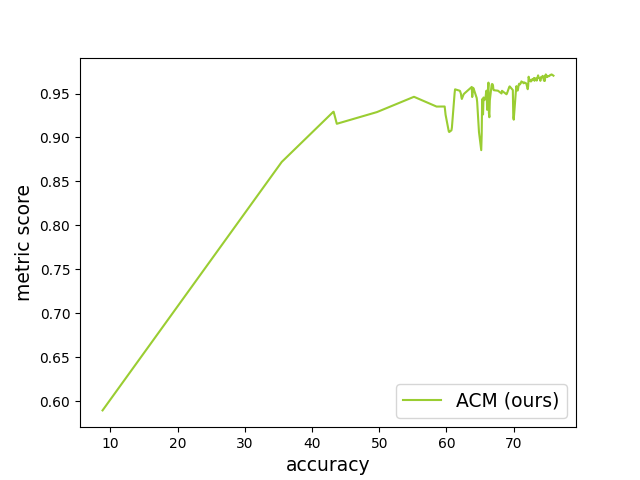}
    }
    \caption{The visualization of the relation between the ACM score and target accuracy. Each sub-figure contains models trained by DANN or CDAN methods on DomainNet or VisDA2017 datasets.}
    \label{fig:visual_2}
\vspace{-2mm}
\end{figure}

\subsection{Hyperparameter Searching Results}

We show the results of hyper-parameter searching on DomainNet, OfficeHome, and Office31 in Tab~\ref{tab:domainnet-search}, Tab~\ref{tab:officehome-search}, and Tab~\ref{tab:office31-search}. The hyper-parameters found by our ACM metric outperform the default hyper-parameters for all four training methods.

\section{Visualizations}

We visualize the consistency between the metric score of our ACM and target accuracy and get some sense of Pearson's correlation between them. In Fig.~\ref{fig:visual_1} and Fig.~\ref{fig:visual_2}, we plot the metric score according to the target accuracy of the models trained by the DANN (CDAN) method on various datasets.

As we can see from the figures, it is clear that the ACM score is positively related to target accuracy. Therefore, when the target accuracy increases, the ACM score tends to increase. This correlation is especially obvious in OfficeHome and DomainNet datasets. 

\section{Limitations and Future Works}

Although we have studied various UDA metrics and proposed new metrics for UDA evaluation, the best derivation of the best model (``dev'') remains 1\%-2\%. It is desirable to propose a new metric that better meets the three criteria of robust metrics. One possible direction is combining multiple metrics to evaluate the model. Meanwhile, the time cost of evaluating the metric should also be considered, and metrics in the paper require, at most, to train a simple network. 
In the paper, we use a simple TPE searcher and relatively small search spaces. More advanced searching strategies and neural architecture searching~\cite{NAS} for UDA can be explored. The paper mainly focuses on the single-source UDA for close-set classification. The unsupervised metric for more transfer learning scenarios can be studied, e.g., Partial UDA, Source-Free UDA, UDA for object detection, semantic segmentation, and depth estimation.

\end{document}